\documentclass[11pt, a4paper, goog]{google}

\usepackage[authoryear, sort&compress, round]{natbib}
\uselogo{} 

\title{Training Optimal Large Diffusion Language Models}

\renewcommand{\today}{}

\definecolor{blue1}{rgb}{0.02352941176,0,1}
\definecolor{jinjie_color}{rgb}{0.43, 0.43, 0.88}

\DeclareMathOperator*{\argmin}{arg\,min~}

\usepackage{hyperref}
\hypersetup{colorlinks=true, urlcolor=blue, linkcolor=blue, citecolor=blue}

\author{
Jinjie Ni$^1$\thanks{Correspondence to: Jinjie Ni <\texttt{jinjieni@nus.edu.sg}>}, Qian Liu, Chao Du$^2$, Longxu Dou$^2$, Hang Yan$^4$, Zili Wang$^3$,\authorcr Tianyu Pang$^2$, Michael Qizhe Shieh$^1$\\
$^1$National University of Singapore, $^2$Sea AI Lab, $^3$StepFun, $^4$Shanghai Qiji Zhifeng Co., Ltd.
}

\begin{document}

\renewcommand{\absfont}{\linespread{1.2}\fontsize{10}{12}\selectfont}

\newlength{\ablabelwidth}
\setlength{\ablabelwidth}{1.8em} 

\newcommand{\abitem}[2][]{%
  \par\noindent
  \hangafter=1
  \hangindent=\ablabelwidth
  \makebox[\ablabelwidth][l]{\textbullet\hspace{0.5em}}
  \if\relax\detokenize{#1}\relax
    #2%
  \else
    {\bfseries #1} #2%
  \fi
}

\begin{abstract}
\textbf{We introduce \textcolor{blue1}{Quokka}, the first large-scale scaling law for diffusion language models (DLMs), encompassing both compute-constrained and data-constrained regimes, and studying the key modeling and optimization designs. Quokka is a good friend of Chinchilla and provides wider scopes. We hope the results would bring short-term practical guidance in DLMs training and long-term inspirations for the whole AI community. We summarize some takeaways below:} \vspace{0.5em}

\abitem[Compute-constrained scaling law.] With fixed FLOPs $C$, the optimal parameters $N_{\mathrm{opt}}\!\propto\!C^{0.5}$ and data size $D_{\mathrm{opt}}\!\propto\!C^{0.5}$, scaling at the same pace; DLMs are \emph{2--5}$\times$ more data-hungry than autoregressive (AR) models at the same $C$—favor smaller models and larger corpora (Figure \ref{fig:compute_constrained_allocations_ar_and_dlm}). We provide a direct comparison with Chinchilla scaling law coefficients in Table~\ref{tab:scaling_coefficients} and their practical optimal allocation comparisons in Table~\ref{tab:quokka_vs_chinchilla}. \vspace{0.2em}

\abitem[Data-constrained scaling law.] Validation loss is U-shaped in epochs $e$; the onset of overfitting scales roughly as $e_{\mathrm{opt}}\!\propto\!U_D^{0.39}/N^{0.55}$, where $N$ is the model size and $U_D$ is the unique data size; e.g., a 10B model on $1$T unique tokens tolerates $\sim$1{,}100 epochs before degradation. We provide practical allocation guidance in Table~\ref{tab:max_epoch_data_constrained}. \vspace{0.2em}

\abitem[Joint allocation under data constraints.] For a larger unique data size $U_D$, the optimal parameter-epoch allocation uses \emph{modestly larger} $N$ and \emph{more} epochs–both $N_{\mathrm{opt}}$ and $e_{\mathrm{opt}}$ increase with $U_D$. We provide practical allocation guidance in Table~\ref{tab:opt_allocation_data_constrained}. \vspace{0.2em}

\abitem[Masked outperforms uniform diffusion at scale.] Absorbing-mask transitions consistently outperform uniform ones on pretrain loss and downstream metrics (\S \ref{subsec:mask_vs_uniform}). \vspace{0.2em}

\abitem[Schedules and curricula.] A linear $\alpha_t$ schedule is strongest in most cases and most stable; poly2 performs better on some benchmarks; an easy$\to$hard noise curriculum (clean-to-noisy $t$ sampling) accelerates early learning and yields small end-of-training gains (\S \ref{subsec:diffusion_schedules}). \vspace{0.2em}

\abitem[Losses.] MaskGIT loss (no importance sampling) converges faster initially, but the principled diffusion ELBO attains better final performance (\S \ref{subsec:diff_loss_formula}). \vspace{0.2em}

\abitem[Hyperparameters transfer.] Batch-size and learning-rate laws from AR models can be carried over for DLM training (\S \ref{subsec:batch_size_and_lr}). \vspace{0.2em}

\abitem[Weight decay.] Little benefit at one epoch, but useful in long multi-epoch runs and for controlling parameter norms (stability in \texttt{bf16}); keep WD when repeating data heavily (\S \ref{subsec:weight_decay}). \vspace{0.2em}
\end{abstract}

{\renewcommand{\thefootnote}{\textdagger}
\maketitle
}

\renewcommand{\thefootnote}{}
\footnotetext{\textcolor{blue1}{This is an initial draft that will be further improved.}}
\renewcommand{\thefootnote}{\arabic{footnote}}

\newpage

\section{Introduction}

\begin{figure}[t]
\centering
\includegraphics[width=0.9\textwidth]{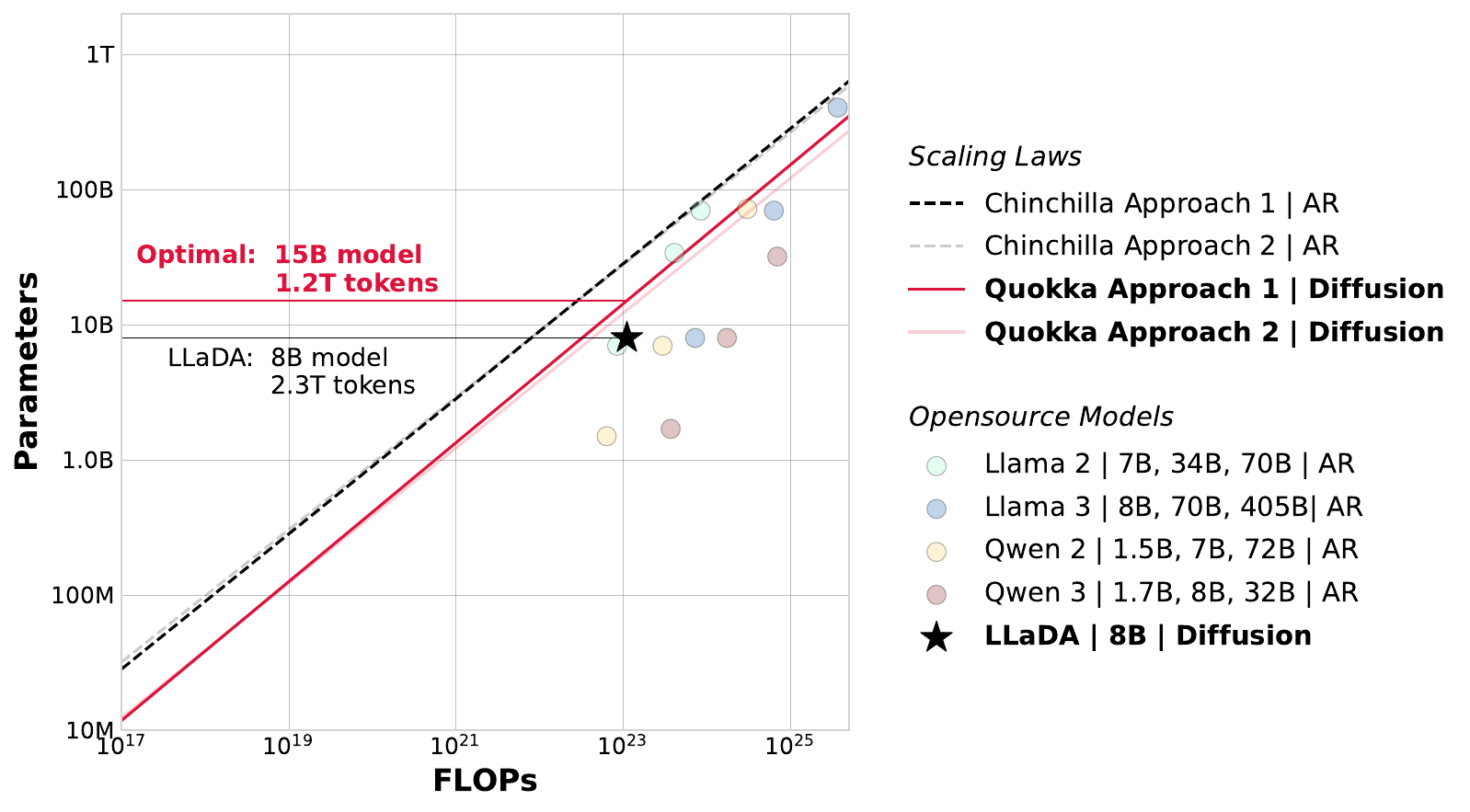}
\caption{\textbf{Overlaid predictions from Chinchilla and Quokka (compute-constrained).} We overlay the predictions from our approach 1 and 2, along with those from \citep{hoffmann2022training}. Though scaling at the same pace, DLMs are \emph{2--5}$\times$ more data-hungry than AR models at the same FLOPs—favor smaller models and larger corpora. We mark the position of LLaDA \citep{nie2025large} in the same space, finding that it's severely over-trained with 2$\times$ smaller models and 2$\times$ more corpora against the Quokka efficient frontier. Meanwhile, wo show the positions of opensource models, finding that most models are over-trained compared with the Chinchilla efficient frontier, except some models from the Llama family. Note that the token statistics are based on the numbers in their reports, which might not be strictly unique tokens. More discussions are detailed in \S \ref{sec:discussions}.}
\label{fig:compute_constrained_allocations_ar_and_dlm}
\end{figure}

2025 marks the first year of diffusion language models (DLMs) scaling. Based on the great efforts that laid the theoretical foundation for DLMs \citep{lou2023discrete,shi2024simplified,ou2024your,sahoo2024simple}, \cite{nie2025large} successfully trained the first large diffusion language model from scratch, competitive to state-of-the-art open-source autoregressive (AR) models \citep{dubey2024llama}. Meanwhile, several commercial DLMs emerged, exhibiting superior coding and math performance with remarkably low generation latency at the same time \citep{deepmind2025geminiDiffusion,khanna2025mercury,song2025seed}. Thereafter, \cite{ni2025difflm} showed that DLMs exhibit much better data learning potential than AR models when data is the bottleneck, a.k.a. "intelligence crossovers", demonstrating a core advantage over AR models under token crisis \citep{xue2023repeat,muennighoff2023scaling}.

DLMs exhibit several modeling advantages over AR models. Their bidirectional attention and diffusion objective enable any-order modeling, allowing data to be modeled in arbitrary directions during both training and inference. This property is particularly beneficial for tasks requiring non-causal dependencies and back-and-forth reasoning, such as coding \citep{xie2025dream,wu2025fast}, mathematics \citep{deepmind2025geminiDiffusion}, report generation \citep{han2025deep}, etc. DLMs' bidirectional attention natively support on-the-fly context modification as new content is generated, a desirable feature in these tasks. Multi-token generation is also natively supported by DLMs, providing the foundation for their bleeding fast decoding. Moreover, DLMs spend more parallelable FLOPs at both the inference and training time, leading to their superior data learning capability and potentially stronger reasoning capabilities. 

However, the knowledge on how to train large DLMs from scratch is still near blank. Existing studies are largely heuristic or simply extrapolate conclusions from AR models \citep{nie2025large,ye2025dream}. In practice, two scaling laws are of primary interest: (1) the compute-constrained (or compute-optimal) scaling law \citep{hoffmann2022training}, where compute is fixed while model and dataset size are unconstrained; and (2) the data-constrained scaling law \citep{muennighoff2023scaling}, where dataset size is fixed while model size and compute are unbounded. Both regimes raise key questions about scaling behavior under these restrictions and, more critically, how to optimally allocate the remaining degrees of freedom. Moreover, beyond the classic trade-offs among data, parameters, and compute, additional modeling and optimization choices can also substantially affect the end-of-training performance of language models.

In this work, we will empirically investigate the dependence of language modeling loss and downstream evaluations on all of these factors. We introduce \textbf{Quokka}, the first large-scale scaling law for DLMs, covering both the compute-constrained and data-constrained regimes, and studying the key modeling and optimization designs. Specifically, the key contributions we made include:

\paragraph{Compute-constrained scaling laws for DLMs.} Under compute-constraint, we revisit the question: Given a fixed FLOPs budget, how should one trade-oﬀ model size and the number of training tokens? To answer this question, we model the final pre-training loss as a function of the number of model parameters $N$, and the number of training tokens, $D$. Since the computational budget $C$ is a deterministic function $FLOPs(N,D)$ of the number of these two variables, our objective is to minimize $L$ subject to the constraint $FLOPs(N, D) = C$:

\begin{equation}
N_{\text{opt}}(C), D_{\text{opt}}(C) = 
\argmin_{N,D \ \text{s.t.} \ \text{FLOPs}(N,D)=C} \ L(N,D).
\end{equation}

The functions $N_{\text{opt}}(C)$ and $D_{\text{opt}}(C)$ characterize the optimal allocation of a compute budget $C$. We estimate these functions empirically using results from a large set of models, spanning parameter counts from under 7M to over 11B and trained on datasets from 1B to over 260B tokens. Across two independent approaches, we consistently find that $N$ and $D$ should scale proportionally with $C$: doubling $N$ requires doubling $D$, mirroring the scaling behavior observed in AR models. Meanwhile, both approaches indicate that DLMs require roughly $2$–$5\times$ more data than AR models under the same FLOPs budget (Figure \ref{fig:compute_constrained_allocations_ar_and_dlm}).

\paragraph{Data-constrained scaling laws for DLMs.}  
In the data-constrained regime—which represents the long-term practical bottleneck—we study the interactions among training performance, unique dataset size, model parameters, and data repetition. We focus on two central questions: (1) Given a fixed model size, a limited amount of unique data, and effectively unlimited compute, how many epochs can the model be trained before performance degradation occurs? (2) Given a fixed unique data budget and unlimited compute, what is the optimal allocation of parameters and data repetitions?  

To address these questions, we model the U-shaped validation loss $L(N, U_D, e)$ as a function of parameters $N$, unique tokens $U_D$, and training epochs (or repetitions) $e$, using results from 21,345 training runs. For question (1), with $N$ and $U_D$ fixed, we seek the maximum number of epochs that minimizes validation loss along the U-curve. FLOPs are excluded from this formulation, as compute is assumed unconstrained:

\begin{equation}
e_{\text{opt}}(\hat{N}, \hat{U}_D) = 
\argmin_{e \ \text{s.t.} \ N=\hat{N},U_D=\hat{U}_D} \ L(e).
\label{eq:dc_question1_intro}
\end{equation}

For question (2), with $U_D$ as the only constraint, we aim to determine the optimal allocation of model size $N$ and epochs $e$. Since performance under data constraints is non-monotonic w.r.t. both $N$ and $e$, the loss surface admits at least one minimum. We therefore fit the joint allocation of $N$ and $e$ that minimizes $L$:

\begin{equation}
e_{\text{opt}}(\hat{U}_D), N_{\text{opt}}(\hat{U}_D) = 
\argmin_{e, N \ \text{s.t.} \ U_D=\hat{U}_D} \ L(e, N).
\label{eq:dc_question2_intro}
\end{equation}

In \S \ref{subsec:opt_model_scaling}, we plot the predicted loss contour $L(N, U_D, e)$, and gave practical suggestions based on the results of Equation~\eqref{eq:dc_question1_intro} and \eqref{eq:dc_question2_intro}. E.g., we can train a 10B model for maximally 1098 epochs on 1T data before seeing a rise in the loss.

\paragraph{Key modeling and optimization designs.}  
Beyond the interplays between parameters, dataset size, data repetition, and compute, we also ablate several critical modeling and optimization choices for DLMs. These include transition kernels (\S \ref{subsec:mask_vs_uniform}), diffusion schedules (\S \ref{subsec:diffusion_schedules}), curriculum strategies (\S \ref{subsec:diffusion_schedules}), loss formulation (\S \ref{subsec:diff_loss_formula}), and optimization hyperparameters such as learning rate (\S \ref{subsec:batch_size_and_lr}), batch size (\S \ref{subsec:batch_size_and_lr}), and weight decay (\S \ref{subsec:weight_decay}). Our results show that while DLMs exhibit markedly different scaling coefficients from AR models, the established AR scaling laws for learning rate and batch size transfer directly.

\section{Preliminaries}
\label{sec:preliminaries}

\subsection{Chinchilla Scaling Law and Its Data-Constrained Version for AR Models}

\paragraph{Chinchilla Scaling Law.}
\cite{hoffmann2022training} studies \emph{compute-constrained} (or compute-optimal) AR pre-training by triangulating evidence from three complementary approaches:
(i) \textbf{Fixed-Parameters}: vary training tokens $D$ while holding model size $N$ fixed;
(ii) \textbf{Fixed-FLOPs (IsoFLOP)}: keep total training compute $C$ fixed while co-varying $N$ and $D$;
(iii) \textbf{Parametric Fit}: fit a two-factor loss surface $L(N,D)$ and derive the compute-optimal allocation.
Its core parametric law is
\begin{equation}
    L(N,D) \triangleq E + \frac{A}{N^{\alpha}} + \frac{B}{D^{\beta}}
    \label{eq:chin}
\end{equation}
with compute $C\!\approx\!6ND$. Minimizing \eqref{eq:chin} at fixed $C$ yields the allocation
\begin{gather}
N_{\mathrm{opt}}(C)=G\left(\frac{C}{6}\right)^{a},\qquad
D_{\mathrm{opt}}(C)=G^{-1}\left(\frac{C}{6}\right)^{b},\\
\text{where}\quad
G = \left(\frac{\alpha A}{\beta B}\right)^{\frac{1}{\alpha+\beta}}, \quad a = \frac{\beta}{\alpha+\beta}, \quad \text{and} \quad b = \frac{\alpha}{\alpha+\beta}.
\label{eq:chin_alloc}
\end{gather}
In practice, $a \approx b$, so compute-optimal training scales $N$ and $D$ in near lockstep. 

\paragraph{A data-constrained generalization.}
When unique data is limited, repeated tokens and excess parameters have \emph{diminishing} marginal value. \cite{muennighoff2023scaling} capture this by replacing the raw $(N,D)$ in Equation~\eqref{eq:chin} by their \emph{effective} counterparts $(N',D')$:
\begin{equation}
L(N,D) \triangleq E + \frac{A}{N'^{\alpha}} + \frac{B}{D'^{\beta}}
\label{eq:dclaw}
\end{equation}
where $D'$ discounts repetitions and $N'$ discounts parameters beyond those needed for the available unique data. Let $U_D=\min\{D,D_C\}$ be the unique tokens used under a data budget $D_C$, and let $R_D=\frac{D}{U_D}-1$ be the number of repeats (epochs beyond the first). Symmetrically, define $U_N$ as the parameters compute-optimal for $U_D$ and $R_N=\frac{N}{U_N}-1$. Then use simple exponential ``half-life'' forms:
\begin{equation}
D' \;=\; U_D \;+\; U_D\,R_D^{*}\!\left(1-e^{-R_D/R_D^{*}}\right),\qquad
N' \;=\; U_N \;+\; U_N\,R_N^{*}\!\left(1-e^{-R_N/R_N^{*}}\right).
\label{eq:effective}
\end{equation}
Here $R_D^{*}$ and $R_N^{*}$ are scale parameters: at $R_D\!=\!R_D^{*}$ (resp.\ $R_N\!=\!R_N^{*}$), each repeated token (resp.\ excess parameter) is worth roughly $(1-1/e)$ of a fresh one. A flaw of this formulation is that it assumes validation loss is non-increasing, which is not true in practice. 

\subsection{Masked Diffusion Language Models}\label{subsec:mdlm}

\textbf{Why masked diffusion?}
DLMs adopt a noising–denoising framework over sequences. Among their variants, \emph{masked diffusion}—also known as absorbing discrete diffusion, which relies on an absorbing transition kernel—has emerged as the most effective formulation \citep{amin2025masking}. It preserves discreteness, supports any-order modeling, enables exact position-wise factorization during corruption, and allows flexible likelihood estimation and natively support multi-token prediction. These properties make masked diffusion a strong competitor to AR modeling while retaining many of its practical advantages. Moreover, \citet{ni2025difflm} demonstrate that masked DLMs consistently outperform AR models under data-constrained regimes through more repetitions on data. This advantage is likely rooted in DLMs’ any-order modeling, high compute-parameter ratio, and inherent data augmentation.

\paragraph{Forward (corruption) process.}
Let $K$ be the vocabulary size, $L$ the sequence length, and $m$ the mask token.
Given a clean sequence $x_0\in\{0,\dots,K{-}1\}^L$, define a monotone diffusion schedule
$\alpha_t\in[0,1]$ with $\alpha_0=1$ and $\alpha_1=0$, where $\alpha_t$ is the probability that a token is \emph{clean} (unmasked) at noise level $t\in[0,1]$.
The forward process independently masks tokens: 
\[
q_{t|0}(x_t\mid x_0)
=\prod_{i=1}^{L} q_{t|0}\!\left(x_t^{(i)}\mid x_0^{(i)}\right),\qquad
q_{t|0}\!\left(x_t^{(i)}\mid x_0^{(i)}\right)
=\begin{cases}
\alpha_t,& x_t^{(i)}=x_0^{(i)},\\[2pt]
1-\alpha_t,& x_t^{(i)}=m~,
\end{cases}
\]
so that the expected unmasked fraction at level $t$ equals $\alpha_t$.

\paragraph{Reverse (denoising) process.}
Starting from the fully masked sequence $x_1$ and a decreasing schedule $1=t_0>t_1>\dots>t_N=0$, the reverse dynamics from $t$ to $s<t$ acts independently across positions:
\[
q_{s|t}\!\left(x_s^{(i)} \mid x_t\right)=
\begin{cases}
1, & x_t^{(i)}\neq m,\; x_s^{(i)}=x_t^{(i)},\\[4pt]
\displaystyle\frac{1-\alpha_s}{\,1-\alpha_t\,}, & x_t^{(i)}=m,\; x_s^{(i)}=m,\\[8pt]
\displaystyle\frac{\alpha_s-\alpha_t}{\,1-\alpha_t\,}\;q_{0|t}\!\left(x_s^{(i)}\mid x_t\right), & x_t^{(i)}=m,\; x_s^{(i)}\in\mathcal{V}\setminus\{m\},\\[8pt]
0, & \text{otherwise.}
\end{cases}
\]
i.e., already-revealed tokens stay fixed; masked tokens either remain masked with probability $\frac{1-\alpha_s}{1-\alpha_t}$ or are revealed by sampling from a \emph{data-prediction} distribution $q_{0|t}(\cdot\mid x_t)$ with probability $\frac{\alpha_s-\alpha_t}{1-\alpha_t}$.
A key \emph{time-agnostic} property \citep{ou2024your} of masked diffusion is that
\[
q_{0|t}\!\left(x_0^{(i)}\mid x_t\right)=p_{\text{data}}\!\left(x_0^{(i)} \,\middle|\, x_t^{\text{UM}}\right),
\]
the conditional distribution of the clean token depends only on the \emph{unmasked} context $x_t^{\text{UM}}$; it does not depend on $t$ beyond which tokens are visible. This allows the denoiser to be parameterized without an explicit time embedding.

\paragraph{Learning objective.}
Let $p_\theta\!\left(x_0^{(i)}\mid x_t\right)$ approximate $p_{\text{data}}\!\left(x_0^{(i)}\mid x_t^{\text{UM}}\right)$.
Masked diffusion maximizes a variational bound on $\log p_\theta(x_0)$, which can be written as minimizing
\begin{equation}
\mathcal{L} \;=\; \int_{0}^{1} w(t;\alpha)\;
\mathbb{E}_{q_{t|0}(x_t\mid x_0)}\!\left[
\sum_{i:\,x_t^{(i)}=m} -\log p_\theta\!\left(x_0^{(i)}\mid x_t\right)
\right]\mathrm{d}t,
\label{eq:mdm_loss}
\end{equation}
where the importance weight $w(t;\alpha)$ depends only on the schedule and, up to a constant factor, takes the natural form
\[
w(t;\alpha)\;=\;\frac{\alpha'_t}{\alpha_t-1}\,.
\]
Intuitively, $w(t;\alpha)$ compensates for the varying expected number of masked positions across noise levels. For the widely used linear schedule $\alpha_t=1-t$, this reduces to the familiar integrand weight $w(t)=1/t$.

\section{Compute-Constrained Scaling Law for Diffusion Language Models}
\label{sec:compute_constrained_scaling_law}

Constrained compute in model training is inevitable—every player in the AGI race faces limited compute budgets while having effectively unlimited model variants to explore. We therefore ask: \emph{Given a fixed FLOPs budget, how should one optimally trade off model size against the number of training tokens?} Following \cite{hoffmann2022training}, we model the DLM training loss, model size, and dataset size using power-law relationships under the limited-compute, infinite-data regime, where each model is trained for a single epoch.

We present two approaches to address this question. First, we conduct extensive IsoFLOPs runs across a range of compute budgets, varying model sizes up to 11B parameters and dataset sizes up to 260B tokens. This allows us to trace the efficient frontier for compute-optimal allocation between model size and dataset size. Second, we fit the power-law loss function to the final training losses obtained from these IsoFLOPs runs. Both approaches converge on the same conclusion: model size and dataset size should scale proportionally with training compute, i.e., doubling $N$ requires doubling $D$, consistent with findings for AR models. However, both approaches also suggest a substantially higher fixed data allocation—roughly $2\text{--}5\times$ that of AR models—for a given FLOPs budget, implying that DLMs are more data-hungry when trained for only a single epoch. Note that \cite{ni2025difflm} shows that DLMs achieve higher data potential under multi-epoch training.

\subsection{Approach 1: IsoFLOPs Profiles}
\label{subsec:isoflops}

\begin{figure}[t]
\centering
\includegraphics[width=1\textwidth]{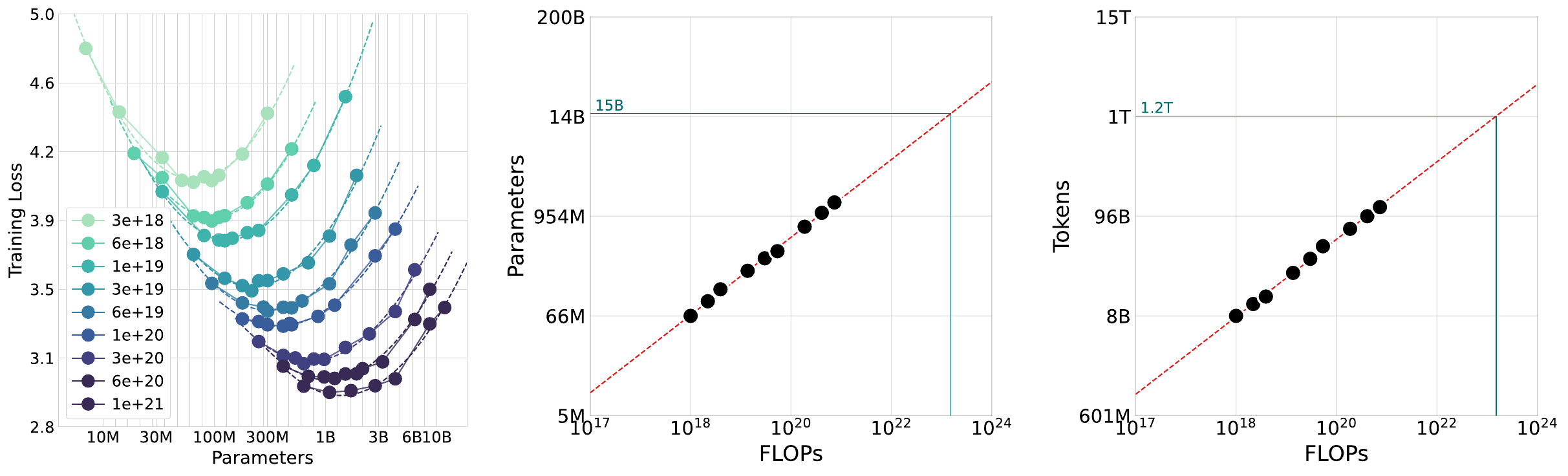}
\caption{\textbf{IsoFLOP curves illustrating the final training loss for a fixed compute budget.} For each curve, we vary the model size and adjust the number of training tokens to maintain constant total training FLOPs. The left panel reveals a distinct performance valley, indicating an optimal trade-off between model size and data for a given compute budget. Leveraging the minima of these curves, we extrapolate the scaling law for the optimal number of parameters and training tokens to larger compute regimes (center and right). The green point highlights our projection for an optimally-scaled model trained with the LLaDA compute budget.}
\label{fig:isoflops_valley_and_linear_fit}
\end{figure}

In the first approach, we vary model size across nine fixed training FLOPs budgets, ranging from $3\times10^{18}$ to $1\times10^{21}$ FLOPs, and record the final training loss at each point. This directly answers the question: for a given FLOPs budget, what is the compute-optimal parameter count?

For each FLOPs budget, we plot the smoothed final loss against parameter count in Figure~\ref{fig:isoflops_valley_and_linear_fit} (left). In all cases, we train a sufficiently diverse set of model sizes to ensure the loss curve exhibits a clear minimum. We fit a parabola to each IsoFLOPs curve to estimate the parameter count at which the minimum loss occurs (Figure~\ref{fig:isoflops_valley_and_linear_fit}, left). We then fit power laws relating compute to the loss-optimal model size and dataset size (Figure~\ref{fig:isoflops_valley_and_linear_fit}, center and right), both of which show near-perfect linearity in log-log space. The resulting scaling exponents are $N_{\text{opt}} \propto C^{a}$ and $D_{\text{opt}} \propto C^{b}$, with $a = 0.51$ and $b = 0.49$. The fitted formulas are $N \approx 0.0216C^{0.514}$ and $D \approx 7.7C^{0.486}$, as summarized in Table~\ref{tab:scaling_coefficients}.

An instructive head-to-head comparison is that the only dense diffusion language model trained from scratch, LLaDA \citep{nie2025large}, which consumed $1.1\times10^{23}$ FLOPs, adopted a suboptimal parameter–data allocation. As shown in Figure~\ref{fig:isoflops_valley_and_linear_fit} (center, right), the compute-optimal allocation at this budget would be a 15B-parameter model trained on 1.2T tokens, rather than the 8B model with 2.3T tokens they used. We provide a direct comparison between AR models and DLMs under compute constraints in \S\ref{sec:opt_model_scaling_compute_constrained}.

\subsection{Approach 2: Fitting a Parametric Loss Function}
\label{sec:parametric_fitting_compute_constrained}

\begin{figure}[t]
\centering
\includegraphics[width=1\textwidth]{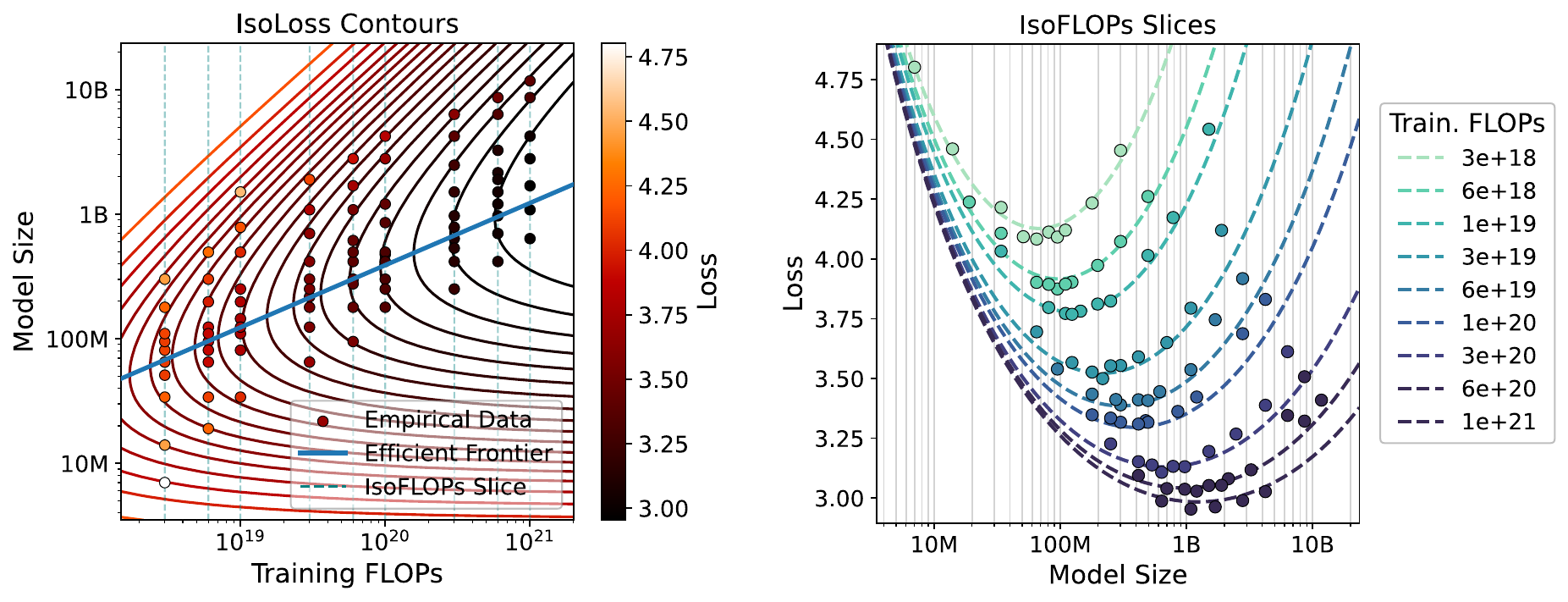}
\caption{\textbf{Parametric fit of the loss function $L(N, D)$.} Left: Iso-loss contours of our fitted model. The blue line indicates the efficient frontier—the trajectory of minimal compute (FLOPs) required to achieve a given loss value, which is linear in log-log space. Right: Several isoFLOPs cross-sections of the loss surface, corresponding to the dashed lines in the left panel. The real data points are also plotted for a comparison.}
\label{fig:parametric_fit_countours_isoslices}
\end{figure}

The second approach models final training loss as a parametric function of model size $N$ (parameter count) and dataset size $D$. Following \cite{hoffmann2022training}, we adopt a functional form based on classical risk decomposition, expressing the loss $L(N,D)$ as:

\begin{equation}
    L(N,D) \triangleq E + \frac{A}{N^{\alpha}} + \frac{B}{D^{\beta}}
\label{eq:compute_constrained}
\end{equation}

This formulation decomposes total loss into three components:  
\begin{enumerate}
    \item \textbf{Irreducible Error ($E$):} The entropy of the true data-generating process, representing the theoretical lower bound on loss, unattainable by any model.  
    \item \textbf{Model Error ($\tfrac{A}{N^{\alpha}}$):} Error due to limited model capacity. Even with infinite data, a finite transformer cannot perfectly capture the true distribution. This term decays as model size $N$ increases.  
    \item \textbf{Training Error ($\tfrac{B}{D^{\beta}}$):} Error from finite dataset size $D$. It captures the gap between a finitely trained model and its fully converged counterpart, diminishing as $D$ grows.  
\end{enumerate}

To estimate the five free parameters $(A, B, E, \alpha, \beta)$, we regress the functional form against our experimental results. Concretely, we minimize the Huber loss \citep{huber1992robust} between predicted and observed log-losses using the L-BFGS algorithm \citep{nocedal1980updating}:  
\begin{equation}
    \min_{A,B,E,\alpha,\beta} \sum_{\text{Runs } i} \text{Huber}_{\delta}\left(\log L(N_i, D_i) - \log L_i^{\text{obs}}\right),
\end{equation}
where $L_i^{\text{obs}}$ denotes the observed loss for run $i$. Log-loss is standard for fitting power-law relationships. We set $\delta = 10^{-3}$ to enhance robustness to outliers, improving predictive accuracy on held-out data. To avoid convergence to poor local minima, we perform a grid search over initial parameter values and retain the fit with the lowest objective value.

A key application of this parametric model is to derive the compute-optimal allocation of a fixed budget $C$ between model size and dataset size. Assuming compute cost scales as $\text{FLOPs}(N, D) \approx 6ND = C$, the optimal $N_{\text{opt}}$ and $D_{\text{opt}}$ are obtained by minimizing Equation~\eqref{eq:compute_constrained} under this constraint. The solution balances model error against training error, yielding a closed-form expression in which both $N_{\text{opt}}$ and $D_{\text{opt}}$ scale as power laws of $C$:
\begin{equation}
    N_{\text{opt}}(C) = G\left(\frac{C}{6}\right)^a, \quad D_{\text{opt}}(C) = G^{-1}\left(\frac{C}{6}\right)^b,
\end{equation}
where the scaling exponents $a$ and $b$, and the constant $G$, are functions of the fitted parameters from our loss model:
$$
G = \left(\frac{\alpha A}{\beta B}\right)^{\frac{1}{\alpha+\beta}}, \quad a = \frac{\beta}{\alpha+\beta}, \quad \text{and} \quad b = \frac{\alpha}{\alpha+\beta}.
$$

By construction, $a + b = 1$. The contours of the fitted loss function $\hat{L}$ and the corresponding efficient frontier are shown in Figure~\ref{fig:parametric_fit_countours_isoslices} (left); Figure~\ref{fig:parametric_fit_countours_isoslices} (right) shows several isoFLOPs cross-sections of
the loss surface, corresponding to the dashed lines in the left panel, with the real data points for a comparison. Our empirical fit, summarized in Table~\ref{tab:scaling_coefficients}, yields exponents $a \approx 0.50$ and $b \approx 0.50$, suggesting that under a fixed compute budget, training data scales at the same pace of parameters. This outcome is fully consistent with approach~1, reinforcing the robustness of the conclusion. From approach~2, the fitted form of Equation~\eqref{eq:compute_constrained} is:

\begin{equation}
    L(N,D) \approx 2.413 + \frac{798.6}{N^{0.379}} + \frac{4604.9}{D^{0.378}}
\end{equation}

\subsection{Optimal Model Scaling}
\label{sec:opt_model_scaling_compute_constrained}

\begin{table}[t]
\centering
\caption{\textbf{A comparison of scaling law coefficients between our model (Quokka) and Chinchilla.} Both DLMs and AR models exhibit similar scaling exponents, implying that the optimal model size and number of training tokens scale at a similar rate. However, for a compute-optimal configuration, our findings suggest allocating $2.2-6.7\times$ more training data with a correspondingly smaller model than prescribed by Chinchilla. We also observe that DLMs have a higher irreducible loss.}
\label{tab:scaling_coefficients}
\begin{tabular}{rrrrrr}
\toprule
Approach              & E    & a    & b    & $k_N$  & $k_D$  \\ \midrule
Chinchilla Approach 1 & -    & 0.50 & 0.50 & 0.09 & 1.88 \\
Chinchilla Approach 2 & -    & 0.49 & 0.51 & 0.15 & 1.15 \\
Chinchilla Approach 3 & 1.69 & 0.46 & 0.54 & 0.60 & 0.28 \\
\textbf{Quokka Approach 1}     & -    & 0.51 & 0.49 & 0.02 & 7.70 \\
\textbf{Quokka Approach 2}     & 2.41 & 0.50 & 0.50 & 0.04 & 4.10 \\ \bottomrule
\end{tabular}
\end{table}

As detailed above, the optimal parameter count $N_{\text{opt}}$ and token budget $D_{\text{opt}}$ follow a power-law relationship with compute $C$: $N_{\text{opt}} \propto C^{a}$, $D_{\text{opt}} \propto C^{b}$. Introducing multipliers $k_N$ and $k_D$, we write $N_{\text{opt}} = k_N C^{a}$ and $D_{\text{opt}} = k_D C^{b}$. 

Table~\ref{tab:scaling_coefficients} summarizes the fitted coefficients and compares them directly with Chinchilla scaling. Despite methodological differences, both approaches of Quokka yield consistent exponents $a$ and $b$, suggesting that model size and training data should scale nearly proportionally with compute. However, while the exponents align, the multipliers $k_N$ and $k_D$ differ, and these dominate the actual allocation under fixed $C$ when $a \approx b$. Since $C = 6ND$, the constraint $k_N \times k_D = \tfrac{1}{6}$ holds. 

Empirically, Quokka exhibits a $2.2$–$6.7\times$ larger $k_D$ than Chinchilla, implying substantially more data and correspondingly fewer parameters are optimal at fixed compute. In practice, for very large FLOPs budgets, even small exponent differences (e.g., $0.51$ vs.\ $0.50$) become increasingly important, eventually outweighing multiplier effects (Table~\ref{tab:quokka_vs_chinchilla}).

DLMs also exhibit a higher irreducible loss than AR models (2.41 vs.\ 1.69). This is intuitive: beyond the intrinsic noise in real-world data, diffusion LMs optimize a variational upper bound (ELBO) on the negative log-likelihood. The forward noising process, discretization, and parameterization introduce a non-vanishing variational gap, so even at infinite scale the extrapolated irreducible loss under the diffusion objective remains higher than that of AR models trained directly on NLL.  

Note that \cite{hoffmann2022training} employed three fitting methods. We merge their approaches 1 and 2 into Quokka approach 1, as they are effectively equivalent. Their approach 3, in contrast, reported negative curvature in the $N \rightarrow N_{\text{opt}}$ frontier, yielding lower $N_{\text{opt}}$ estimates. Accordingly, for coefficients other than the irreducible loss $E$, we compare against Chinchilla approaches 1 and 2. The irreducible loss is reported only under their approach 3, i.e., the parametric fit.

\begin{table}[]
\centering
\caption{\textbf{Optimal FLOPs and training tokens allocation for compute-optimal models.} For a range of model sizes, we plot the estimated training FLOPs and number of tokens required to achieve compute optimal, as predicted by Approach 1, to provide a practical guidance for DLMs training. The estimates for both approach 1 and 2 are close, presented in Table~\ref{tab:quokka_approach1n2}. We also included the numbers predicted by Chinchilla scaling law to perform a head-to-head comparison.}
\label{tab:quokka_vs_chinchilla}
\begin{tabular}{r|rrrr}
\toprule
\multicolumn{1}{c}{} & \multicolumn{2}{c}{\textbf{Quokka}} & \multicolumn{2}{c}{Chinchilla} \\ \midrule
Parameters           & FLOPs         & Tokens     & FLOPs          & Tokens        \\ \midrule
400 M                 & 9.46e+19      & 39.3 B      & 1.92e+19       & 8.0 B         \\
1 B                   & 5.62e+20      & 93.5 B      & 1.21e+20       & 20.2 B        \\
10 B                  & 4.96e+22      & 825.2 B     & 1.23e+22       & 205.1 B       \\
67 B                  & 2.01e+24      & 5.0 T       & 5.76e+23       & 1.5 T         \\
175 B                 & 1.30e+25      & 12.4 T      & 3.85e+24       & 3.7 T         \\
280 B                 & 3.24e+25      & 19.3 T      & 9.90e+24       & 5.9 T         \\
520 B                 & 1.08e+26      & 34.6 T      & 3.43e+25       & 11.0 T        \\
1 T                   & 3.86e+26      & 64.2 T      & 1.27e+26       & 21.2 T        \\
10 T                  & 3.41e+28      & 566.4 T     & 1.30e+28       & 216.2 T       \\ \bottomrule
\end{tabular}
\end{table}

Table~\ref{tab:quokka_vs_chinchilla} reports the estimated FLOPs and token counts required for models of different sizes to lie on the compute-optimal frontier, alongside Chinchilla’s allocations. Across scales, DLMs consistently allocate $2$–$5\times$ more tokens than AR models. This follows naturally: under compute constraints, data is not the bottleneck, so each example is used once. Unlike AR models, DLMs require corruption of inputs during training, effectively demanding more data to represent the same amount of information. As a result, DLMs favor comparatively smaller models trained on substantially larger corpora. These findings offer practical guidance for pre-training DLMs in compute-limited regimes.

\section{Data-Constrained Scaling Law for Diffusion Language Models}
\label{sec:data_constrained_scaling_law}

In the long run, compute will not be the bottleneck in the pursuit of greater intelligence. According to Common Crawl's official statistics \citep{CommonCrawl2025CrawlSize}, web data grows roughly linearly, whereas compute for training AI models grows exponentially \citep{SevillaRoldan2024ComputeGrowth}. Since compute can be scaled both by increasing chip counts and by extending training time, it is effectively unbounded. By contrast, data constitutes the true limiting factor. In particular, certain domains face acute scarcity, including non-English language data, high-quality code, mathematical text, medical data, etc.

\begin{figure}[t]
\centering
\includegraphics[width=1\textwidth]{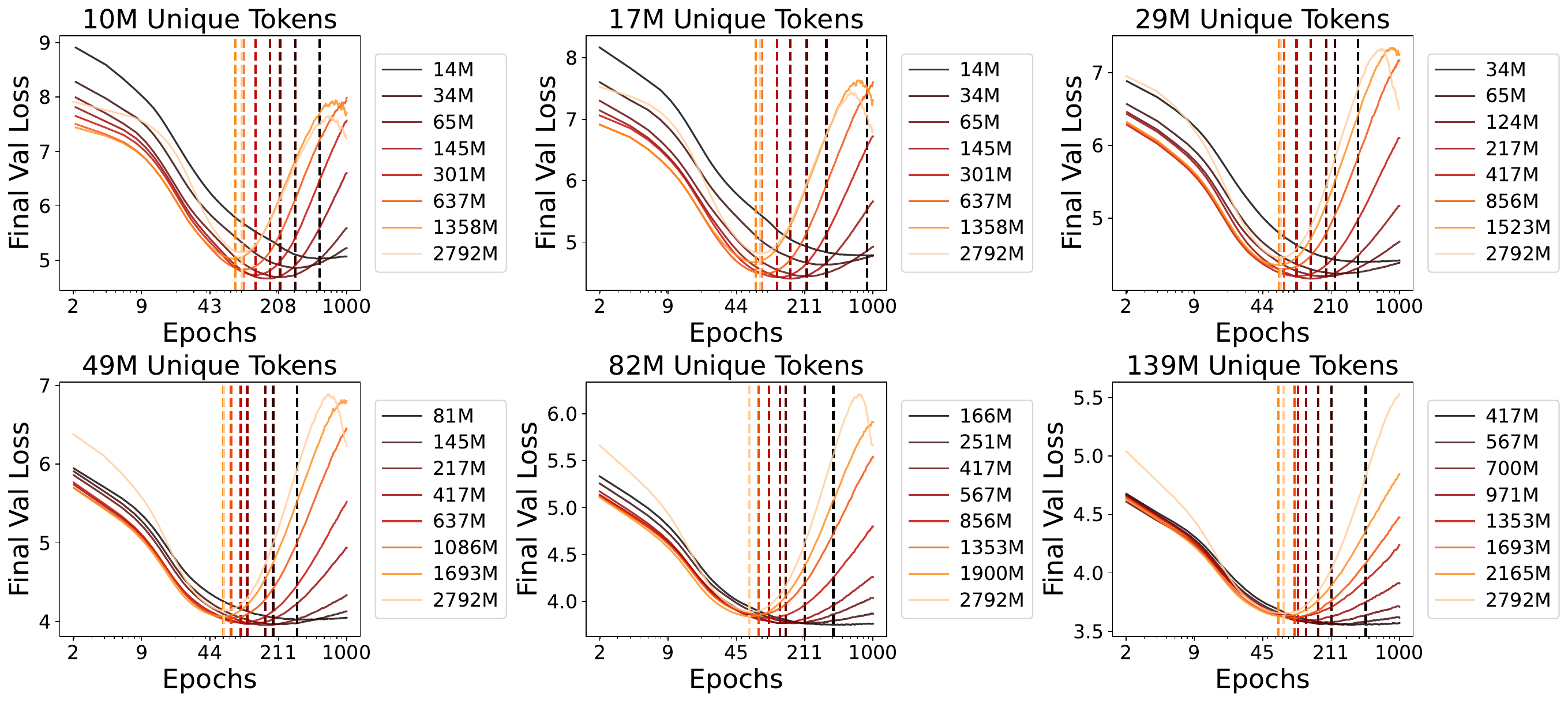}
\caption{\textbf{Final-step validation losses for models of varying sizes trained with different unique data budgets and epochs.} We consistently observe a U-shaped relationship between model size and final validation loss for a fixed data budget, with a minority of runs exhibiting double descent. Larger model sizes tend to accelerate the onset of overfitting (the right side of the "U"), while increasing the number of unique tokens delays it. The minimum achievable loss improves as the amount of unique data increases. These empirical findings provide the motivation for our data-constrained scaling law.}
\label{fig:combined_loss_vs_epochs}
\end{figure}

Under data constraints, a practical approach to improving model performance is repeated data usage, such as multi-epoch training. Our primary goal is to quantify the effect of multi-epoch training on performance and its relationship with unique dataset size and model parameter allocation. We address this by modeling the loss landscape with respect to training epochs $e$, model parameters $N$, and unique dataset size $U$. Beyond this, we focus on two key questions:

\begin{itemize}
    \item \emph{Given a fixed model size, a fixed unique-data budget, and unbounded compute, how many epochs can we train before performance degrades?}\vspace{0.5em}
    \item \emph{Given a fixed unique-data budget and unbounded compute, can we predict the optimal allocation between model size and number of training epochs?}
\end{itemize}

\subsection{An Effort in Modeling the Validation Loss with Overfitting} 

Modeling validation loss is substantially more challenging than modeling pre-training loss. \citet{muennighoff2023scaling} proposed Equation~\eqref{eq:dclaw} to capture validation loss, introducing the notion of diminishing "effective model size" and "effective data size," which reflects the intuition that repeated exposure to the same data yields diminishing performance gains. However, this formulation has a critical flaw: it produces a monotonically non-increasing validation loss, which contradicts reality. In practice, repeated training on the same data inevitably leads to increased validation loss due to overfitting, a direct consequence of the bias–variance tradeoff.  

To better characterize the validation loss landscape, we trained a suite of DLMs across varying parameter scales, unique data sizes, and epochs—amounting to 24,000 runs (Figure~\ref{fig:combined_loss_vs_epochs}). The results clearly demonstrate the onset of overfitting when training on limited data for extended periods. At the same time, these experiments reveal several intriguing patterns that informed the design of our proposed formulation:

\begin{itemize}
    \item For any model size and unique data budget, validation loss eventually increases once trained for sufficiently many epochs.
    \item With a fixed unique data budget, smaller models overfit more slowly.
    \item With fixed model size, larger unique data budgets delay overfitting.
    \item The minimum achievable loss decreases monotonically with unique data size.
    \item For a fixed unique data size, the minimum achievable loss is non-monotonic w.r.t. model size: it first decreases as capacity grows, then increases as overfitting dominates.
\end{itemize}

With that in mind, we proposed the below formula:

\begin{equation}
\begin{aligned}
L(N, U_D, e) \triangleq E + \frac{A}{N^{\alpha}} + \frac{B}{D'^{\beta}} \\\\
\text{where} \quad D' &= U_D \cdot e^{p_{e}} \cdot \exp\left(-\left(\frac{\max(0, e - 1)}{e_{p}}\right)^{\gamma}\right)
 \quad \text{and} \quad e_{p} &= c_{p} \frac{U_D^{m_{p}}}{N^{k_{p}}}
\end{aligned}
\end{equation}

Our formulation introduces ten coefficients to fit: the irreducible loss $E$, and the parameters $\alpha$, $\beta$, $A$, $B$, $c_{p}$, $m_p$, $k_p$, $p_e$, and $\gamma$. This functional form extends the Chinchilla scaling law to capture the U-shaped validation loss curves characteristic of multi-epoch training under data constraints. The key modification is replacing the dataset size $D$ in Chinchilla with an "effective dataset size" $D'$, which depends on the number of epochs $e$, model size $N$, and unique data size $U_D$. This formulation has the following desirable properties:

\textbf{Full learning–overfitting cycle modeling.} The effective dataset size $D'$ is defined as the product of a learning term ($e^{p_e}$) and an overfitting penalty ($\exp(\dots)$). At small $e$, the learning term dominates, $D'$ increases, and validation loss decreases. At large $e$, the penalty dominates, $D'$ shrinks, and validation loss rises—capturing the complete learning–overfitting cycle and aligning with the first observation.

\textbf{Capturing the dynamics of the optimal epoch.} The peak overfitting epoch $e_p$ explicitly models the trade-off between model size and data budget. The numerator term $U_D^{m_p}$ ensures that more unique data postpones overfitting (larger $e_p$), while the denominator term $N^{k_p}$ reflects that larger models overfit more quickly (smaller $e_p$). This directly accounts for the second and third observations.

\textbf{Predicting optimal performance limits.} The formulation preserves the core structure of a scaling law. A larger unique data budget $U_D$ increases the attainable peak of $D'$, yielding a lower minimum validation loss, consistent with the fourth observation. For the fifth observation, the interaction between the capacity term ($A/N^\alpha$) and the data–overfitting term ($B/D'^\beta$, with $D'$ dependent on $N$ via $e_p$) reproduces the U-shaped dependence of optimal loss on model size under a fixed data budget.

\textbf{Natural reduction to the compute-constrained law when $e \le 1$.} A key property of this formulation is that it generalizes compute-constrained formula in a consistent way. At one epoch of training ($e \le 1$), the $\max(0, e-1)$ term vanishes, the exponential penalty equals 1, and the effective dataset size reduces to $D' = U_D \cdot 1^{p_e} = U_D$. The loss then simplifies to $L = E + A/N^\alpha + B/U_D^\beta$, exactly recovering the compute-constrained law for a model of size $N$ trained on $U_D$ tokens.  

By adding only five parameters ($c_p, m_p, k_p, p_e, \gamma$) to the original five compute-constrained coefficients, the formulation effectively models the three-dimensional optimization space of model size, unique data budget, and training epochs.

\begin{figure}[t]
\centering
\vspace{-0.2cm}
\includegraphics[width=1\textwidth]{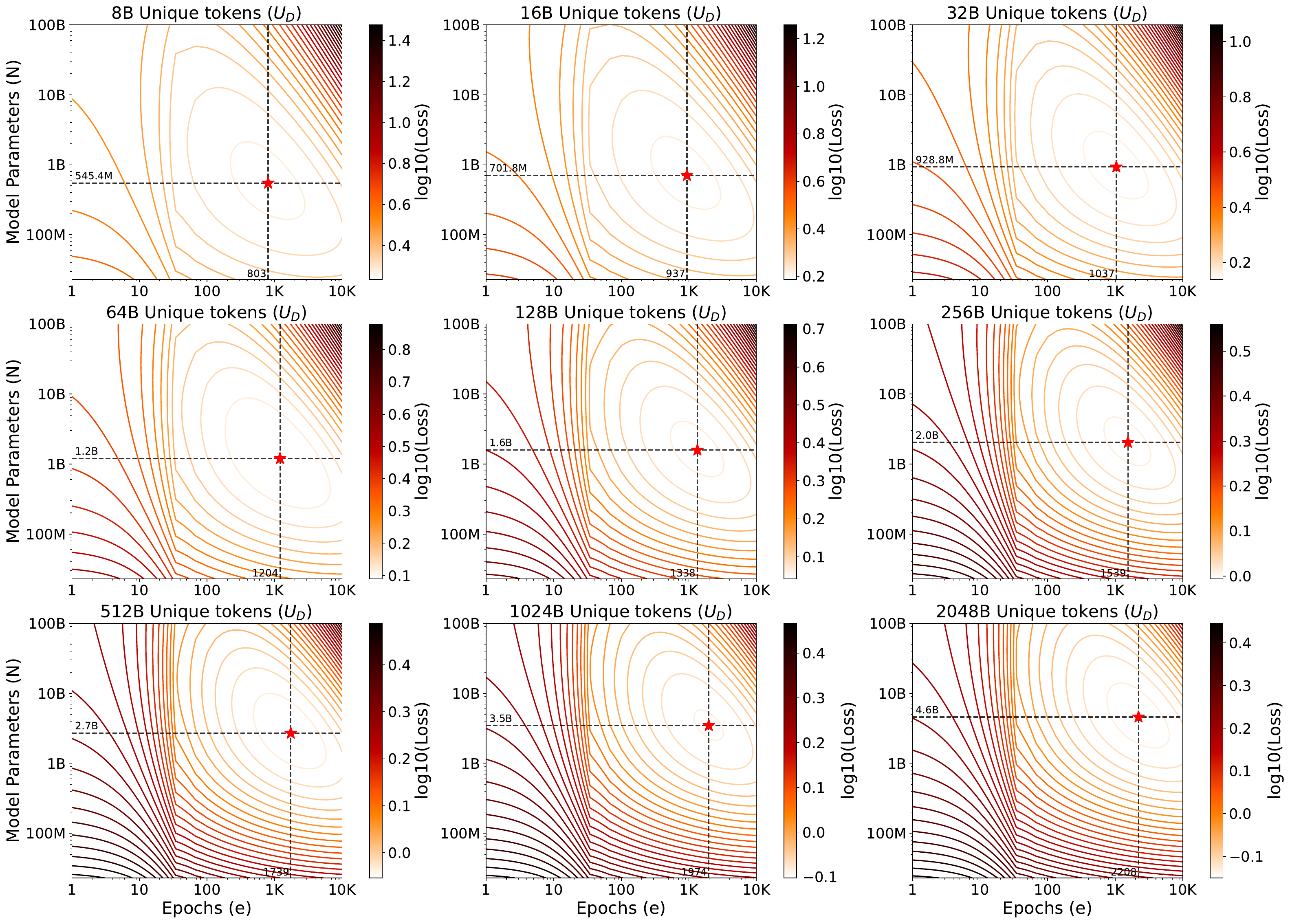}
\caption{\textbf{The loss contours predicted by the fitted data-constrained loss $\hat{L}(N, U_D, e)$.} We exhibit the $N$ - $U_D$ contours with different unique data budgets $U_D$. We observe a local optima within each observation scope and the optimal $N$ and $e$ consistently grow with $e$.}
\label{fig:data_constrained_countour}
\end{figure}

\subsection{Optimal Model Scaling}
\label{subsec:opt_model_scaling}

We fit the proposed formula on 23,145 runs spanning different values of $N$, $U_D$, and $e$, using the same fitting procedure as in the compute-constrained setting. The resulting fitted form is:

\begin{equation}
\begin{aligned}
L(N, U_D, e) &= \frac{1535.23}{N^{0.42}} + \frac{54.21}{\left(U_D \cdot e^{1.49} \cdot \exp\left(-\left(\frac{\max(0, e - 1)}{254.35 \frac{U_D^{0.39}}{N^{0.55}}}\right)^{0.40}\right)\right)^{0.13}} \\ 
\label{eq:fitted_dclaw}
\end{aligned}
\end{equation}

From the fitted formula, we can interpret that the onset of overfitting scales roughly as $e_{\mathrm{opt}}\!\propto\!U_D^{0.39}/N^{0.55}$. The irreducible loss diminishes to a negligible value and is omitted, likely because the interaction among $N$, $U_D$, and $e$ implicitly induces an effective lower bound. Using the fitted validation loss form, we plot loss contours in the $(N, e)$ plane under varying $U_D$, which predict the validation landscape given $(N, U_D, e)$ and provide guidance for the two central questions. The results reveal the existence of local optima when the unique-token budget is fixed. Moreover, larger unique-token budgets generally require both larger models and more epochs to be fully exploited. However, the extremely low validation losses predicted in the contours may not be fully attainable in practice due to fitting error.

\begin{table}[t]
\centering
\caption{\textbf{The maximum epochs one can train given the model parameters $N$ and unique tokens $U_D$}, predicted by the fitted data-constrained loss function \ref{eq:fitted_dclaw}, answering question 1.}
\label{tab:max_epoch_data_constrained}
\begin{tabular}{r|rrrrrrrrr}
\toprule
$N$ / $U_D$ & 10 M & 100 M & 1 B  & 10 B  & 100 B & 1 T   & 10 T   & 100 T  & 1000 T \\ \midrule
400 M  & 70  & 175  & 430 & 1057 & 2593 & 6357 & 15585 & 38205 & 93651 \\
1 B    & 42  & 105  & 260 & 641  & 1572 & 3857 & 9456  & 23180 & 56821 \\
10 B   & 11  & 29   & 73  & 181  & 447  & 1098 & 2693  & 6603  & 16187 \\
67 B   & 1   & 9    & 25  & 63   & 157  & 388  & 953   & 2339  & 5736  \\
175 B  & 1   & 4    & 14  & 37   & 93   & 229  & 564   & 1385  & 3398  \\
280 B  & 1   & 1    & 10  & 28   & 71   & 177  & 436   & 1072  & 2629  \\
520 B  & 1   & 1    & 7   & 20   & 50   & 126  & 311   & 764   & 1875  \\
1 T    & 1   & 1    & 4   & 13   & 35   & 88   & 217   & 535   & 1312  \\
10 T   & 1   & 1    & 1   & 1    & 9    & 24   & 61    & 151   & 373   \\ \bottomrule
\end{tabular}
\end{table}

\begin{table}[t]
\centering
\caption{\textbf{The optimal model parameters $N$ and epochs $e$ allocation under different unique tokens $U_D$}, predicted by the fitted data-constrained loss function \ref{eq:fitted_dclaw}, answering question 2.}
\label{tab:opt_allocation_data_constrained}
\begin{tabular}{r|rrr}
\toprule
Unique Tokens & Parameters & Epochs & FLOPs    \\ \midrule
10 M           & 41 M        & 247    & 6.07e+17 \\
100 M          & 95 M        & 356    & 2.04e+19 \\
1 B            & 222 M       & 569    & 7.59e+20 \\
10 B           & 518 M       & 910    & 2.83e+22 \\
100 B          & 1.6 B       & 1151   & 1.11e+24 \\
1 T            & 3.7 B       & 1842   & 4.12e+25 \\
10 T           & 8.7 B       & 2947   & 1.54e+27 \\
100 T          & 20.2 B      & 4715   & 5.72e+28 \\
1000 T         & 47.1 B      & 7543   & 2.13e+30 \\ \bottomrule
\end{tabular}
\end{table}

The fitted formula also enables practical guidance for training DLMs under data constraints. Table~\ref{tab:max_epoch_data_constrained} addresses the first question: \emph{given model size, unique data budget, and unbounded compute, how many epochs can be run before performance degradation occurs?} For reference, we include representative model parameter counts aligned with the compute-constrained scaling law. Table~\ref{tab:opt_allocation_data_constrained} addresses the second question: \emph{given a fixed unique data budget and unbounded compute, what is the optimal allocation of model size and training epochs?}

\paragraph{Caveats}  
(1) The validation loss landscape remains poorly understood, and its mathematical form is far from established. We do not have a strict theoretical justification for our formulation, and thus cannot claim it holds universally. For instance, we observed double-descent behavior in a small subset of long-epoch runs. In our fitting, we assume a single descent and truncate the second peak for these cases.  
(2) In Figure~\ref{fig:combined_scaling_law}, we compared actual $N$–$e$ optima across data budgets against the fitted ones. The fitted contours tend to overshoot epochs and underestimate model size. Similarly, in Figure~\ref{fig:data_constrained_fitted_vs_actual}, we show actual vs. predicted validation losses for randomly sampled $(N, U_D, e)$. While Equation~\eqref{eq:fitted_dclaw} captures the overall loss shape, noticeable gaps remain in some cases. \S \ref{sec:data_constrained_alternative_formulas_and_fitting} provides alternative formulations and predictions that may also be plausible but resulted in higher loss in the fitting.  
(3) Validation loss values depend heavily on the choice of validation set and tokenizer, making absolute values less meaningful. The emphasis should instead be on trends and the dynamic interplay among variables.

\section{Key Modeling and Optimization Choices}
\label{sec:key_design_choices_and_hps}

Training optimal diffusion language models depends on more than parameter allocation, dataset size, and training epochs. Here, we ablate additional factors. Given resource constraints, a full ablation is infeasible; instead, we focus on the factors we consider most critical.

We report benchmark results on HellaSwag (commonsense reasoning) and MMLU (knowledge), chosen for their popularity and stability across model configurations \citep{liu2023llm360,muennighoff2024olmoe}. Their broad adoption allows direct comparison with prior work, making them reliable indicators for assessing the impact of our ablations.

\subsection{Masked vs. Uniform Transition Kernel}
\label{subsec:mask_vs_uniform}

\begin{figure}[ht]
\centering
\includegraphics[width=1\textwidth]{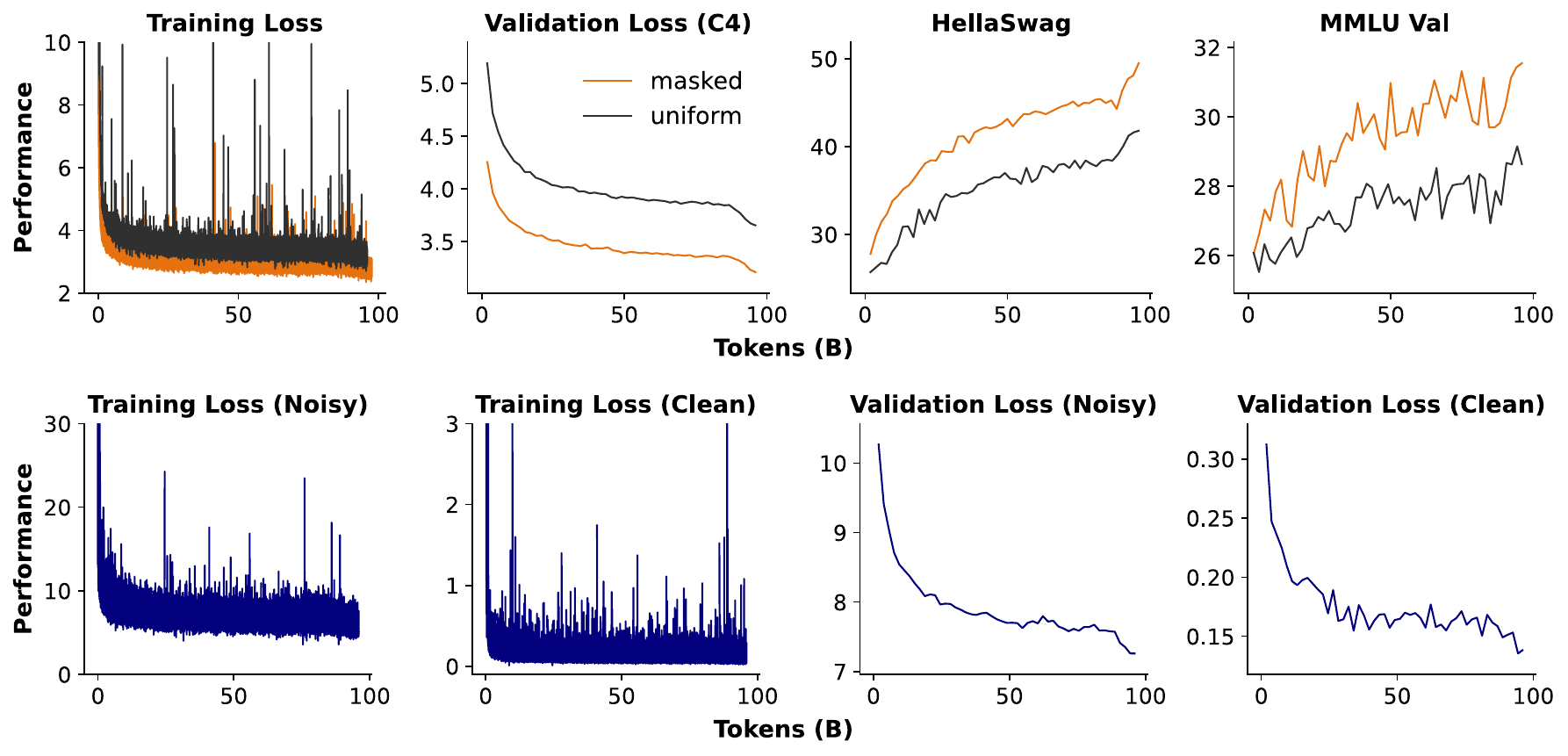}
\caption{Upper: \textbf{Masked and uniform transition kernels.} 1B models are trained on 96B unique tokens. Masked DLMs significantly outperforms the uniform ones. Lower: \textbf{The training and validation loss for the clean and corrupted positions of the uniform DLM.}}
\label{fig:mask_vs_uniform}
\end{figure}

In this ablation, we examine two key variants of discrete diffusion models for language: uniform diffusion and masked (absorbing) diffusion. Their primary difference lies in the state transition rules, which govern how text is corrupted in the forward process and reconstructed in the reverse process.

The \textbf{uniform diffusion} model corrupts a sentence by progressively replacing tokens with randomly sampled ones from the vocabulary, eventually reducing the text to uniform noise. Its reverse process learns to denoise this sequence, gradually refining random tokens into a coherent sentence. In contrast, the \textbf{masked diffusion} model corrupts text by replacing tokens with a special \texttt{[MASK]} token. Its reverse process resembles a fill-in-the-blank task, predicting the original words within a fully or partially masked sequence.

Formally, both dynamics are defined by a continuous-time Markov process with transition rate matrix $Q$. In uniform diffusion, transitions from any token to any other occur at a constant rate, yielding the following $N \times N$ matrix for vocabulary size $N$:
\begin{equation}
Q^{\text{uniform}} =
\begin{pmatrix}
1-N & 1 & \cdots & 1 \\
1 & 1-N & \cdots & 1 \\
\vdots & \vdots & \ddots & \vdots \\
1 & 1 & \cdots & 1-N
\end{pmatrix}
\end{equation}

Here, the off-diagonal entries denote the uniform transition rate to any other token, while the diagonal entries capture the rate of leaving the current token state.

In contrast, masked diffusion restricts transitions to a single absorbing \texttt{[MASK]} state. The rate matrix is structured to enforce this one-way corruption in the forward process while simultaneously defining the generative dynamics for the reverse process. Assuming the final index corresponds to the \texttt{[MASK]} token, the matrix takes the form:
\begin{equation}
Q^{\text{absorb}} =
\begin{pmatrix}
-1 & 0 & \cdots & 0 & 0 \\
0 & -1 & \cdots & 0 & 0 \\
\vdots & \vdots & \ddots & \vdots & \vdots \\
0 & 0 & \cdots & -1 & 0 \\
1 & 1 & \cdots & 1 & 0
\end{pmatrix}
\end{equation}
The diagonal $-1$s specify the transition rate from any token to the \texttt{[MASK]} state. In the forward process, this drives sequences to become fully masked within finite time. The final row of ones encodes the reverse transitions, allowing the \texttt{[MASK]} state to generate any vocabulary token, which the model learns to parameterize.

Masked diffusion is often easier to model, as the task reduces to filling in masked positions rather than distinguishing noise from clean tokens. Both theoretical and empirical studies suggest that masked diffusion models generally outperform uniform ones \citep{amin2025masking, lou2023discrete}. However, direct large-scale comparisons under LLM evaluation settings remain absent.

We compare masked and uniform diffusion pre-training using a broad set of metrics. As shown in Figure~\ref{fig:mask_vs_uniform}, masked diffusion consistently outperforms uniform diffusion across all metrics by a wide margin. To further probe the uniform variants, we plot their losses on both clean and noisy positions. Although the model is not explicitly given indicators for these positions, it learns to distinguish most noisy from clean tokens with low loss (around 0.15). This suggests that the main challenge lies not in identifying noisy versus clean tokens, but in transforming arbitrary embeddings into the correct ones.

It is worth noting that the uniform diffusion loss used here does not compute an exact ELBO, as multiple variants exist and complicate head-to-head comparisons. Instead, we adopt the same reweighting scheme as masked diffusion for noisy positions and average the loss over clean ones, enabling a direct comparison of the learning difficulty between masking and uniform transitions. Additionally, uniform diffusion includes a continuous-time embedding layer, which introduces minimal parameter overhead.

\subsection{Diffusion Schedules}
\label{subsec:diffusion_schedules}

\begin{figure}[ht]
\centering
\includegraphics[width=1\textwidth]{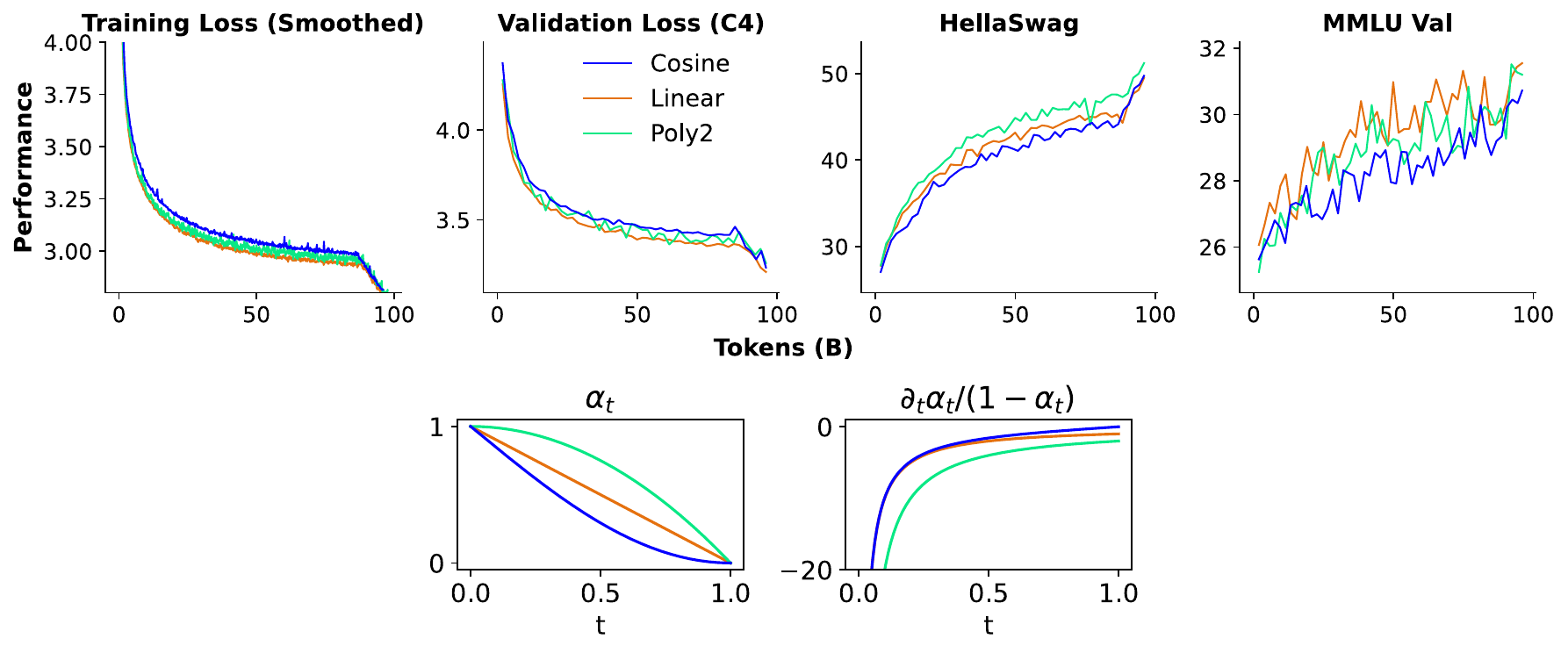}
\caption{Upper: \textbf{Three commonly used diffusion schedules and their performances.} 1B models are trained on 96B unique tokens. Lower: \textbf{The shapes of $\alpha_t$ and cross entropy reweighting $\partial_t \alpha_t / (1 - \alpha_t)$.}}
\label{fig:diffusion_schedule1}
\end{figure}

The diffusion schedule is a central design choice in training DLMs. We ablate two types of schedules. The first is the standard diffusion schedule, a predefined sequence controlling the rate and manner of noise injection and removal at each step. The second is the noise-level sampling schedule across the training lifecycle, where different noise levels are sampled at different stages.

\paragraph{Diffusion schedule.} We examine three common schedules: linear, poly2, and cosine, defined as $\alpha = 1 - t$, $\alpha = 1 - t^2$, and $\alpha = 1 - \cos\!\left(\tfrac{\pi}{2}(1 - t)\right)$, respectively. Figure~\ref{fig:diffusion_schedule1} (bottom) illustrates their shapes. Cosine assigns higher-than-average probability to masking, poly2 lower, and linear lies in between. From pre-training and evaluation results (Figure~\ref{fig:diffusion_schedule1}), cosine performs worst across metrics, while linear consistently outperforms the others in both train/val loss and MMLU. Linear also exhibits lower variance than nonlinear schedules, consistent with \cite{shi2024simplified}. The poly2 schedule achieved better performance on HellaSwag.

\begin{figure}[ht]
\centering
\includegraphics[width=1\textwidth]{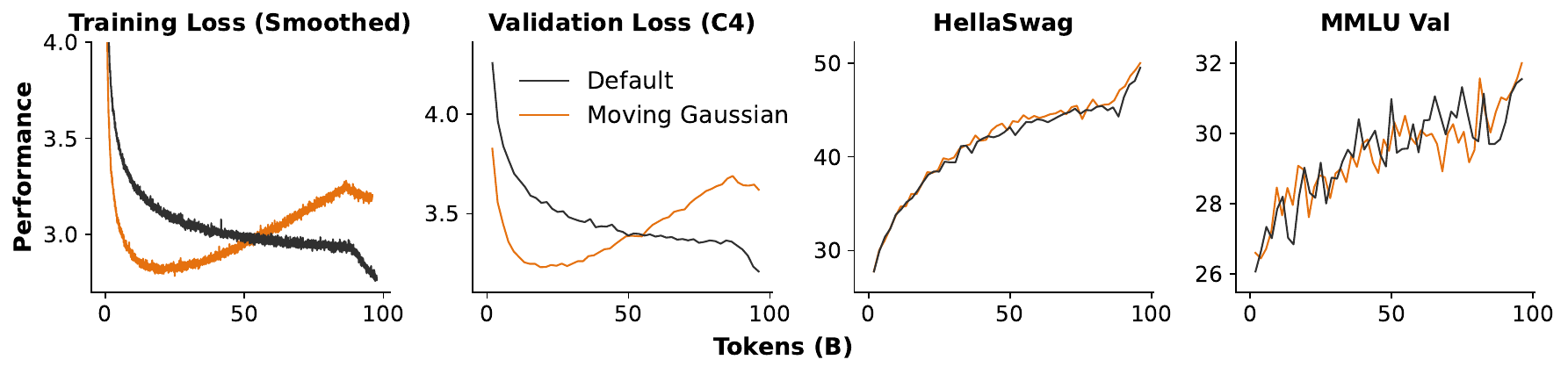}
\caption{\textbf{Uniform $t$ vs. clean-to-noisy $t$ sampling}, where a moving Gaussian window gradually shifts from low-noise sampling early in training to high-noise sampling later, implementing an easy-to-hard curriculum. 1B models are trained on 96B unique tokens.}
\label{fig:diffusion_schedule2}
\end{figure}

\paragraph{Training-time noise schedule.} A natural intuition in training DLMs is to begin with cleaner data and gradually increase noise, aiming for stronger end-of-training performance \citep{zhu2025skyladder}. This is straightforward to implement by adjusting the sampling of $t$. In our setup, we use a moving Gaussian window to bias $t$ toward lower values early in training, so the model first learns easier prediction tasks before progressively transitioning to harder ones as the Gaussian window shifts from 0 to 1. Results show that this schedule yields faster loss reduction in the early stages, followed by rising loss as noisier samples dominate. It achieves slightly better end-of-training performance across both benchmarks, suggesting this direction merits further study.

\subsection{Diffusion Loss Formula}
\label{subsec:diff_loss_formula}

\begin{figure}[ht]
\centering
\includegraphics[width=0.8\textwidth]{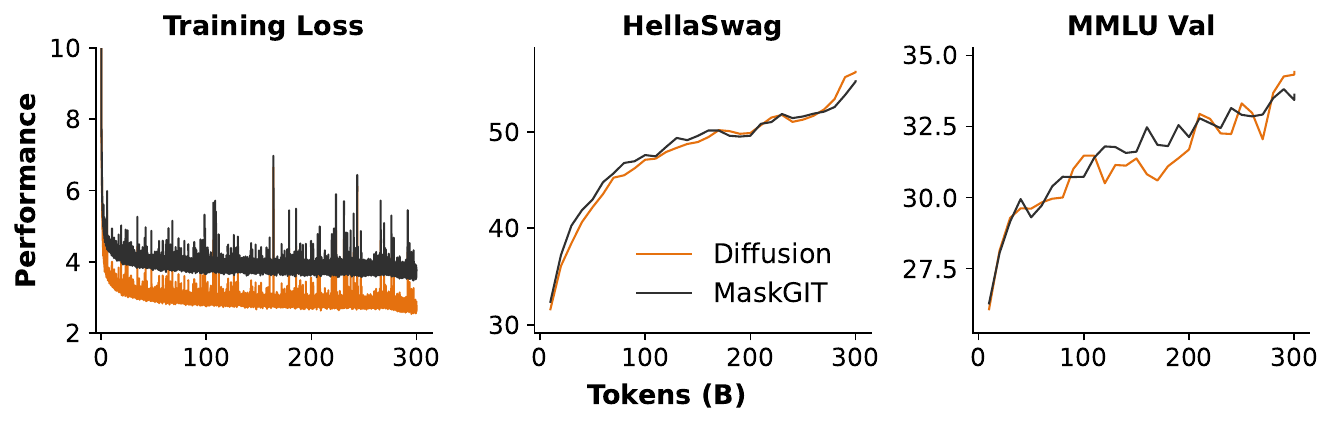}
\caption{\textbf{Principled diffusion loss (Equation~\eqref{eq:mdm_loss}) and MaskGIT loss (Equation~\eqref{eq:mdm_loss} without reweighting).}  1B models are trained on 300B unique tokens.}
\label{fig:diffusion_loss_formula}
\end{figure}

Generative masked language models can be trained using either the principled diffusion loss \citep{shi2024simplified} or the masked loss \citep{chang2022maskgit}. The diffusion loss is generally regarded as more faithful, since it optimizes a likelihood lower bound and is expected to yield better results. We compare masked generative models trained with diffusion loss (Equation~\eqref{eq:mdm_loss}) and MaskGIT loss (Equation~\eqref{eq:mdm_loss} without reweighting) over 300B tokens. Surprisingly, despite not optimizing a principled ELBO, MaskGIT achieves consistently comparable performance throughout training and even converges faster on both evaluations. While diffusion loss ultimately delivers stronger end-of-training performance on both benchmarks, this finding highlights the need for further study of how theoretical bounds influence training dynamics.

\subsection{Batch Size and Learning Rate Transferability}
\label{subsec:batch_size_and_lr}

\begin{figure}[ht!]
\centering
\includegraphics[width=1\textwidth]{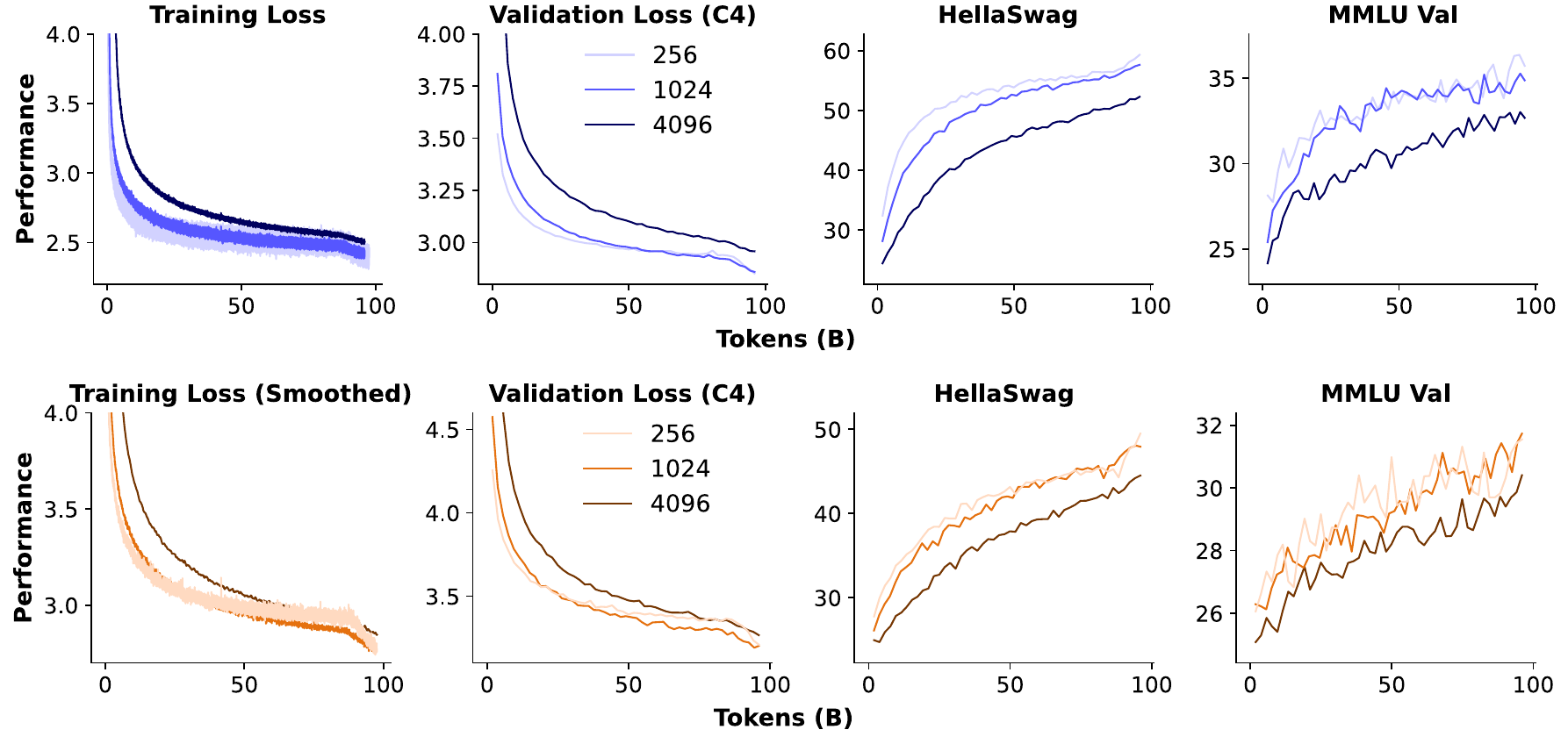}
\caption{\textbf{Batch size transferability from AR models (upper) to DLMs (lower).} Both show consistent trends across batch sizes, suggesting that DLM training can leverage batch size laws from AR studies. 1B models are trained on 96B unique tokens.}
\label{fig:batch_size}
\end{figure}

\begin{figure}[ht!]
\centering
\includegraphics[width=1\textwidth]{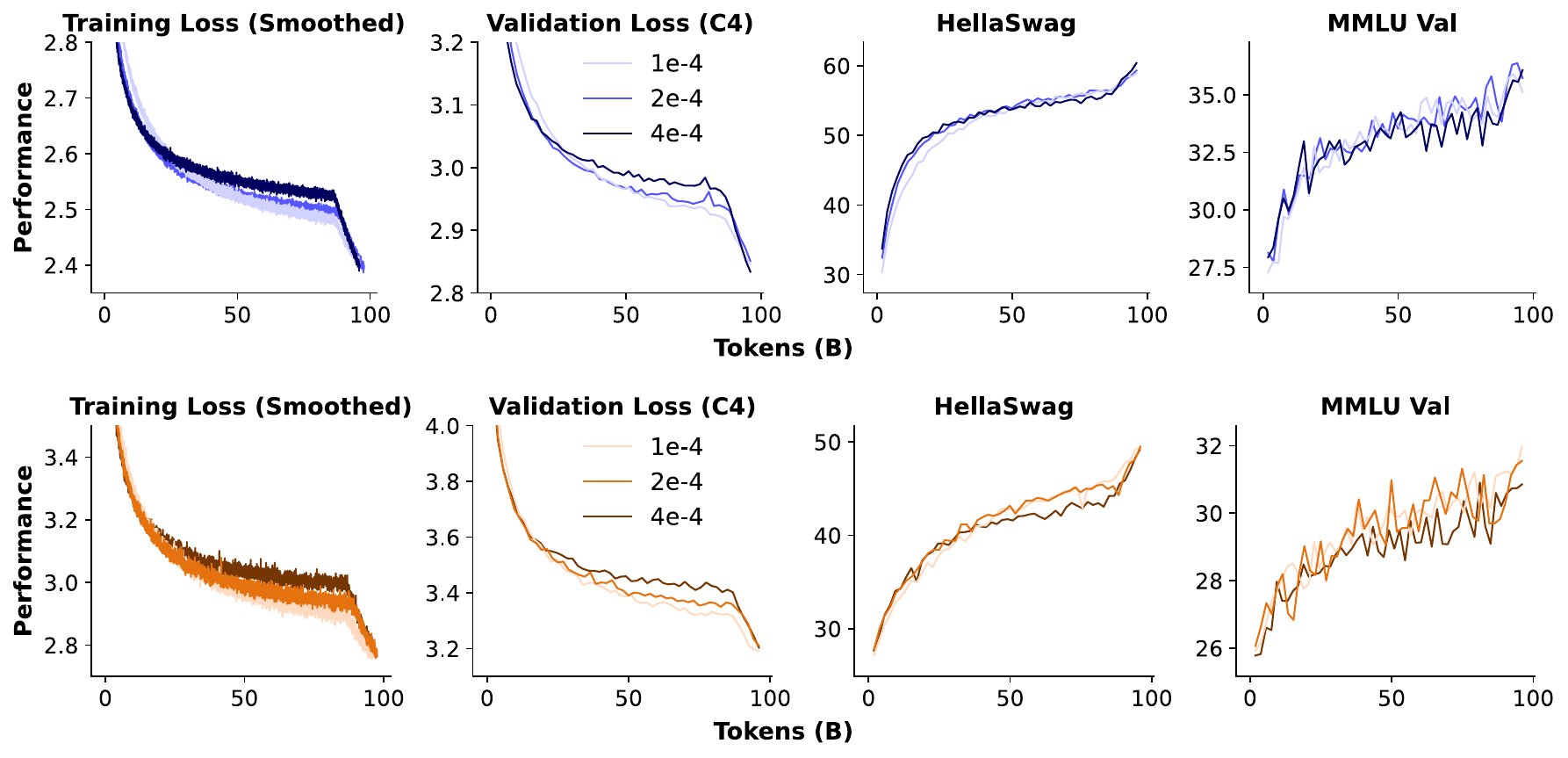}
\caption{\textbf{Learning rate transferability from AR models (upper) to DLMs (lower).} Both show consistent trends across learning rates, suggesting that DLM training can leverage learning rate laws from AR studies. 1B models are trained on 96B unique tokens.}
\label{fig:learning_rate}
\end{figure}

\paragraph{Batch size.} Training hyperparameters such as batch size are critical for stability and performance. AR models have well-established scaling laws for these settings \citep{li2025predictable}. Batch size is closely tied to dataset size, and diffusion language models (DLMs) effectively augment data through noise injection \citep{ni2025difflm}, often exhibiting higher variance in pre-training loss. Larger batches can mitigate both issues, raising the question of whether DLMs favor larger batch sizes than AR counterparts. Surprisingly, as shown in Figure~\ref{fig:batch_size}, training dynamics for AR and DLMs are similar: batch size 4096 lags behind 256 and 1024, with the latter two performing comparably. This suggests that changing the training objective does not alter the optimal batch size when data and model architecture are fixed, implying that established AR scaling laws might be able to transfer directly to DLMs.

\paragraph{Learning rate.} We also examine the transferability of learning rate, another key hyperparameter. We grid search three peak values ranging from small to large. As shown in Figure~\ref{fig:learning_rate}, both AR and DLM models show minimal differences in end-of-training performance across learning rates after annealing, with $1\text{e-}4$ yielding a slight advantage. Convergence speed differences across learning rates are also consistent between AR and DLMs. These results further support that the training objective does not alter the optimal hyperparameter space, and that DLMs can directly reuse established learning rate practices from AR models.

\subsection{Weight decay}
\label{subsec:weight_decay}
\begin{figure}[ht!]
\centering
\includegraphics[width=1\textwidth]{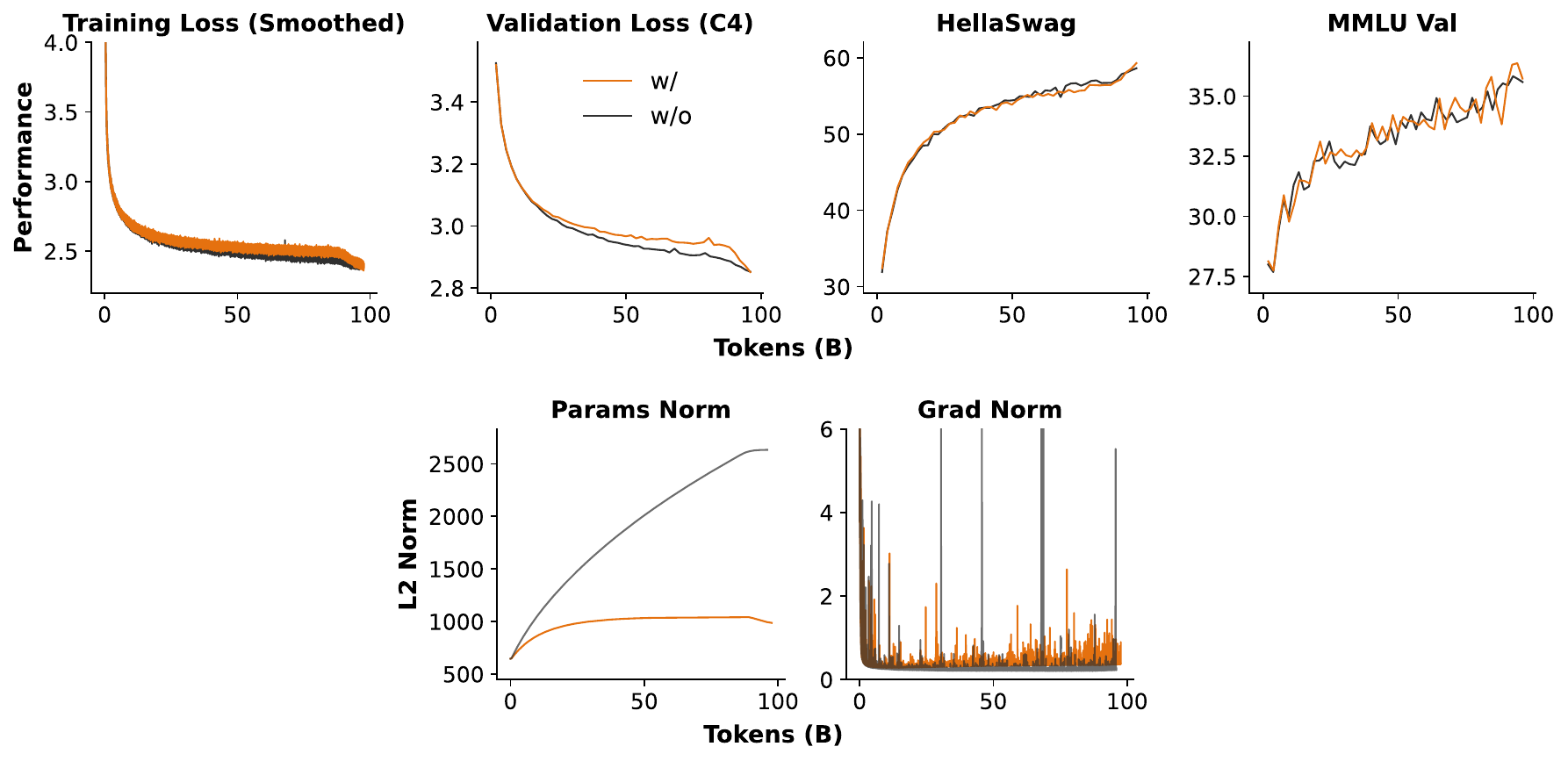}
\caption{\textbf{The impact of weight decay on AR models in single epoch scenarios.} 1B AR models are trained with and without weight decay, on 96B unique tokens.}
\label{fig:weight_decay_96b1e_ar}
\end{figure}

\begin{figure}[ht!]
\centering
\includegraphics[width=1\textwidth]{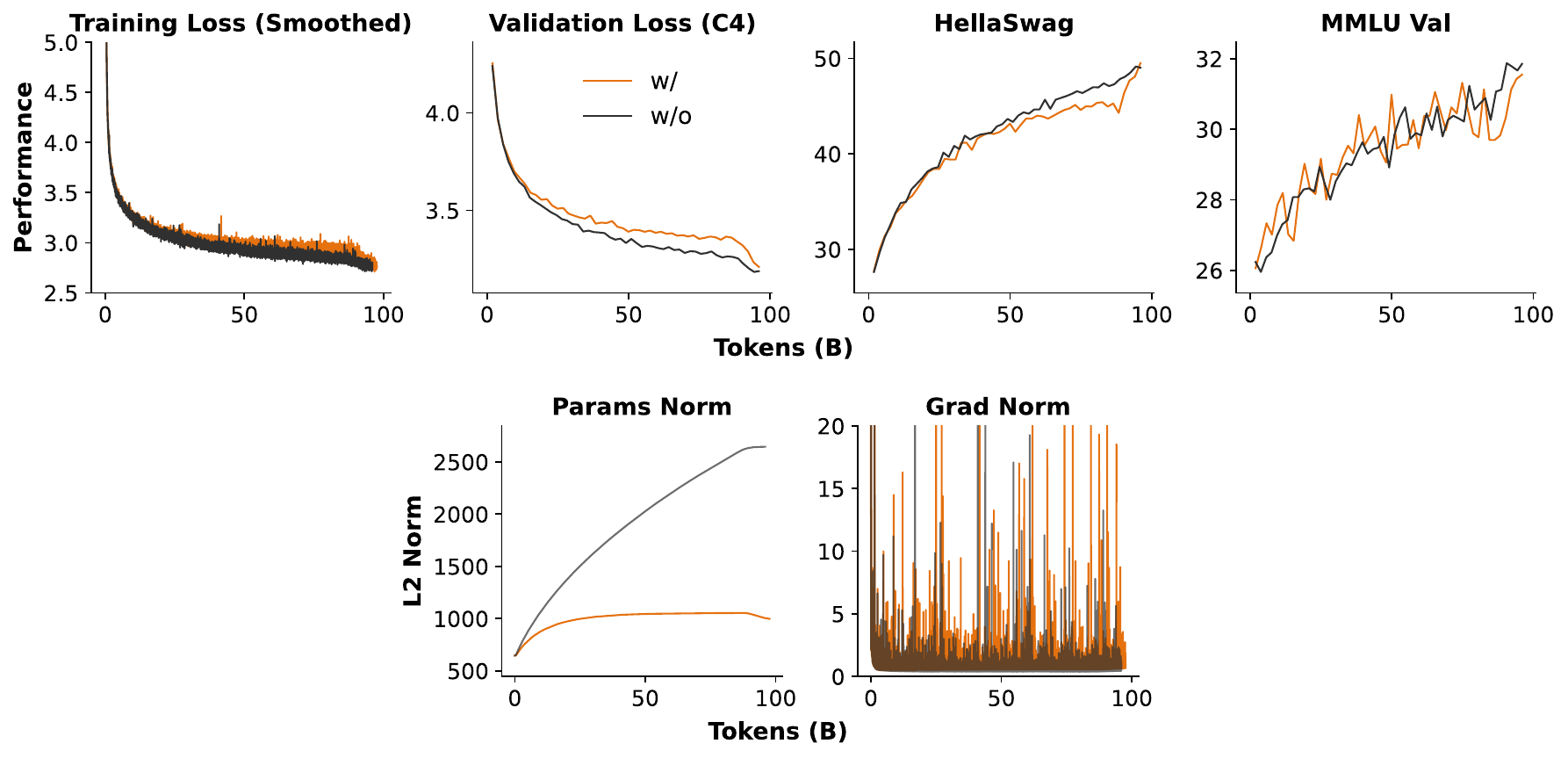}
\caption{\textbf{The impact of weight decay on DLMs in single epoch scenarios.} 1B DLMs are trained with and without weight decay, on 96B unique tokens.}
\label{fig:weight_decay_96b1e_dlm}
\end{figure}

Weight decay is a standard technique in LLM pre-training to keep parameter norms stable and mitigate issues such as overfitting and numerical instability. We investigate its effect in DLM pre-training by comparing AR and DLM models with and without weight decay under two settings: single-epoch and multi-epoch training.

\paragraph{Single-epoch training.} We train models for 96B tokens over 1 epoch. As shown in Figure~\ref{fig:weight_decay_96b1e_ar} and Figure~\ref{fig:weight_decay_96b1e_dlm}, neither AR nor DLM benefits from weight decay in this regime; in fact, removing weight decay leads to faster convergence.

\begin{figure}[ht!]
\centering
\includegraphics[width=1\textwidth]{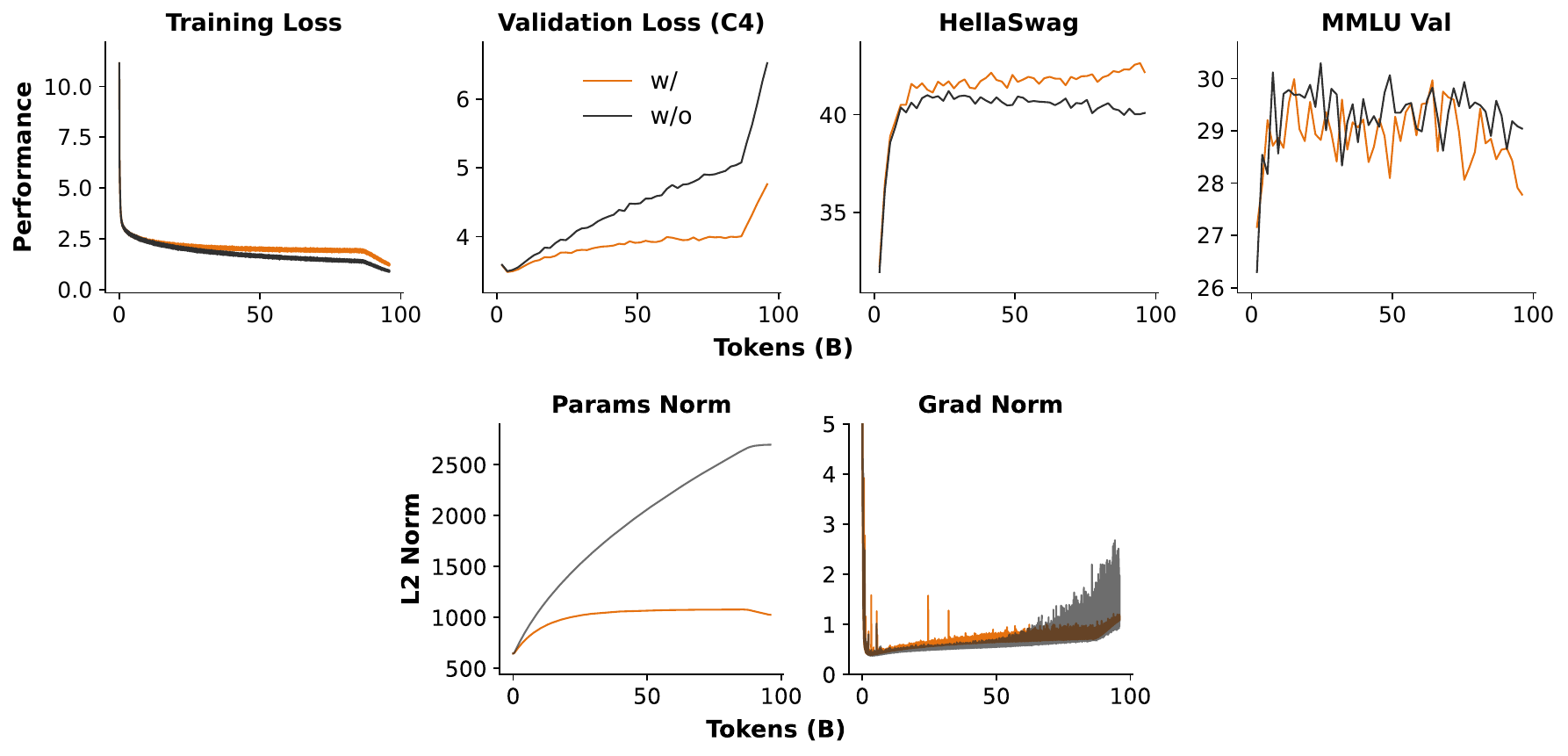}
\caption{\textbf{The impact of weight decay on AR models in multi-epoch scenarios.} 1B AR models are trained with and without weight decay, on 1B unique tokens for 96 epochs.}
\label{fig:weight_decay_1b96e_ar}
\end{figure}

\begin{figure}[ht!]
\centering
\includegraphics[width=1\textwidth]{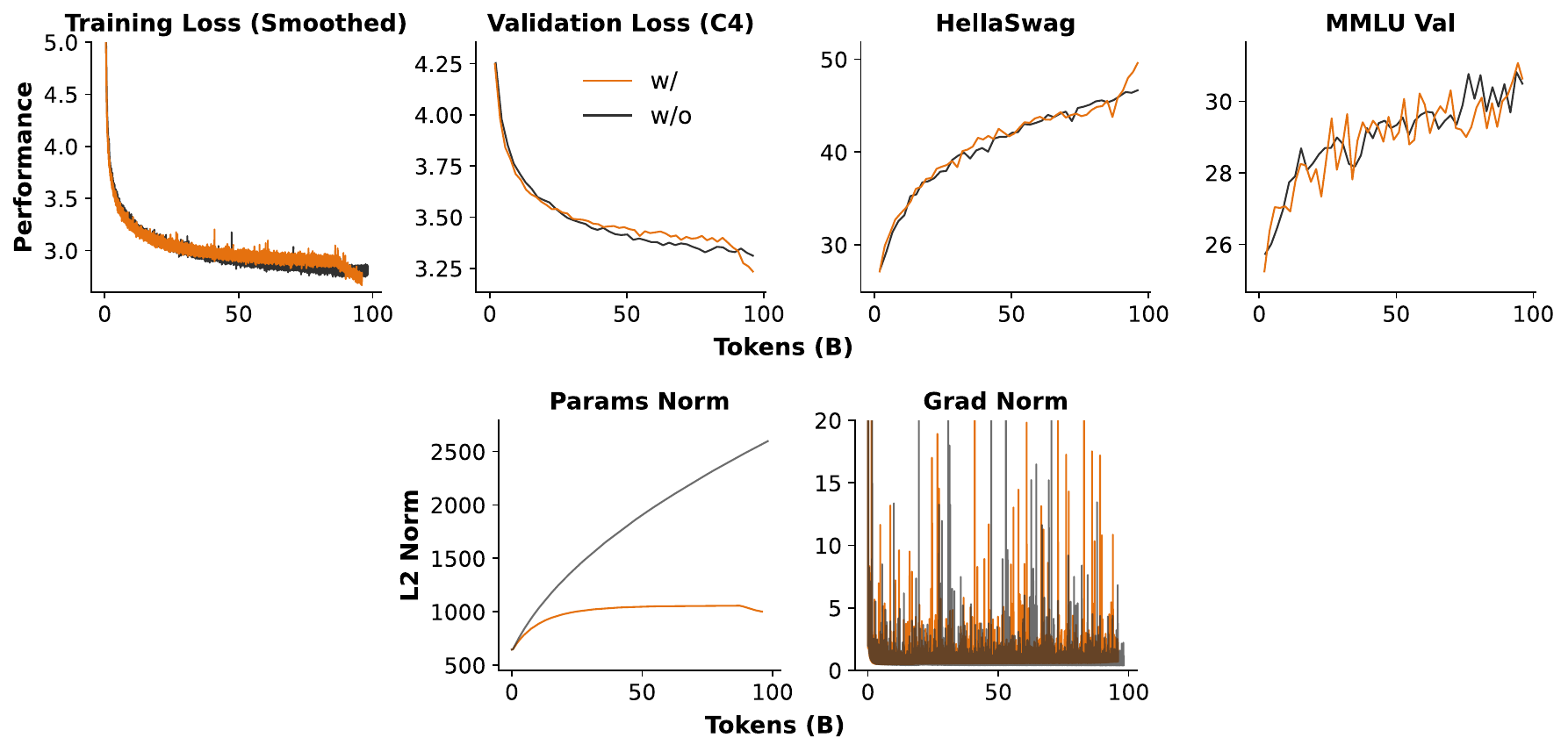}
\caption{\textbf{The impact of weight decay on DLMs in multi-epoch scenarios.} 1B DLMs trained are with and without weight decay, on 1B unique tokens for 96 epochs.}
\label{fig:weight_decay_1b96e_dlm}
\end{figure}

\paragraph{Multi-epoch training.} We train models on 1B unique tokens for 96 epochs, a setting prone to overfitting where weight decay is expected to play a larger role. As shown in Figure~\ref{fig:weight_decay_1b96e_ar}, removing weight decay severely degrades AR models’ validation loss and benchmark performance. In contrast, DLMs remain largely unaffected and appear robust to data repetition even without weight decay. That said, applying weight decay still yields better end-of-training results across all metrics.

Although weight decay shows limited benefit in 3 of 4 ablations, maintaining healthy parameter norms remains important. As shown in Figure~\ref{fig:weight_decay_96b1e_ar}–\ref{fig:weight_decay_1b96e_dlm}, removing weight decay consistently increases the L2 norm of parameters, risking numerical instability. In practice, this can cause logits before softmax to collapse, since \texttt{bf16} provides only 7 mantissas bits and quantization becomes coarse for values above 128, substantially harming both training and inference performance.

\section{Related Work}
\label{sec:related_work}

\subsection{Scaling Laws} 

Understanding how scaling affects large language model (LLM) performance has been a central research focus. The seminal work of \citet{kaplan2020scaling} showed that model performance follows predictable power-law trends with respect to model size, compute, and training data, implying that ever-larger models should yield better results. This paradigm shaped the development of models such as GPT-3 \citep{brown2020language}. However, \citet{hoffmann2022training} challenged this view with Chinchilla, demonstrating that, under a fixed compute budget, optimal performance arises from scaling model size and training data in tandem. This revealed that many prior models, including Gopher \citep{rae2021scaling}, were undertrained, shifting the field’s understanding toward balanced scaling and more efficient compute utilization.

Subsequent research has refined our understanding of scaling laws, particularly in data-constrained regimes. The Chinchilla laws, though influential, assume effectively unlimited training data. As models scale further, the scarcity of unique, high-quality data has emerged as a critical bottleneck. \citet{muennighoff2023scaling} introduced a data-constrained scaling law (Equation~\eqref{eq:dclaw}) to model validation loss under limited data, where repeated exposure reduces the "effective model size" and "effective data size." While this captures diminishing returns, the formulation has a key limitation: it enforces a non-increasing validation loss, whereas in practice repeated epochs inevitably induce overfitting, increasing validation loss due to the bias–variance tradeoff. In this work, we propose a new formulation that addresses this flaw. 

Beyond pre-training loss, scaling law research is expanding to downstream task performance \citep{isik2024scaling}, inference dynamics \citep{wu2025inference}, and theoretical grounding, linking empirical trends to concepts such as data manifold dimensionality \citep{bahri2024explaining, sharma2022scaling}. This broadening scope underscores the need for more refined laws that integrate model architecture, data quality, and task-specific requirements. \citet{li2025predictable} further explored hyperparameter scaling, offering practical guidance for pre-training choices.

For DLMs, systematic scaling laws were lacking prior to Quokka. \citet{nie2024scaling} trained models at low FLOPs budgets to compare scaling trends of AR models and DLMs, marking an important step toward DLM scaling, though their study provided only limited scaling law coefficients \& insights.

\subsection{Diffusion Language Models} 
Building on the theoretical foundations of DLMs \citep{lou2023discrete,shi2024simplified,ou2024your,sahoo2024simple}, \citet{nie2025large} trained the first large-scale DLM from scratch, achieving performance competitive with leading open-source AR models \citep{dubey2024llama}. In parallel, several commercial DLMs have emerged, demonstrating strong coding and math capabilities while offering significantly lower generation latency \citep{deepmind2025geminiDiffusion,khanna2025mercury,song2025seed}. \citet{ni2025difflm} further showed that DLMs possess substantially higher data potential than AR models under limited data, enabling so-called "intelligence crossovers" that highlight their advantage in the face of the token crisis \citep{xue2023repeat,muennighoff2023scaling}. 

Efforts have also explored hybrid approaches bridging AR and diffusion. Block diffusion \citep{arriola2025block} performs block-wise diffusion, with block size 1 similar to AR modeling without shift. Dream \citep{ye2025dream} initialized DLMs with AR priors and employed a "shift-by-one" strategy to better retain AR knowledge, offering another effective training paradigm. Recent work has also advanced DLM coders \citep{gong2025diffucoder,xie2025dream}, DLM RL scaling \citep{zhu2025llada}, accelerated inference techniques \citep{wu2025fast}, pushing DLMs toward greater practicality and competitiveness.

\section{Discussions}
\label{sec:discussions}

In practice, model training is often constrained by resources beyond compute–leading to deviations from the allocations prescribed by scaling laws. For instance, Llama 3 \citep{dubey2024llama} trained an 8B model with 15T tokens, whereas the Chinchilla law would suggest a 70B model for 2T tokens. Several factors contribute to such deviations: (1) Scaling-optimal allocation is not the only consideration for commercial models; factors such as deployability, customer adoption, and hardware compatibility (e.g., GPU/TPU memory limits) play a decisive role. (2) Compute budgets are not always strict. In many cases, one can effectively "expand" compute by extending training time, making smaller models with more data or epochs more practical than adhering rigidly to scaling predictions. (3) Current compute- and data-constrained scaling laws are limited in scope, and their coefficients can shift across architectures and datasets. Thus, scaling laws should be viewed as high-level guidance on balancing model size, data, and training duration, while precise choices require empirical tuning under specific constraints.

\section{Acknowledgment}
We thank Shen Nie, Jiacheng Ye, and Cunxiao Du for their fruitful discussions and pointers.

\bibliography{main}

\appendix

\section{Implementation Details}
\label{sec:imp_details}

All experiments were conducted with a heavily modified Megatron-LM codebase. Compute-constrained runs and ablations were trained on a subset of the Nemotron-CC corpus \citep{su2024nemotron}, while data-constrained runs used a subset of the c4-en corpus \citep{raffel2020exploring}. Validation losses were consistently evaluated on the c4-val split, following \citet{muennighoff2023scaling}. Token budgets were randomly sampled from the respective corpora without additional filtering. Model parameters were initialized from a normal distribution with standard deviation 0.02. Architecturally, we adopted a performant configuration combining the GPT-2 tokenizer, RoPE, SwiGLU, pre-layer RMSNorm, bias-free layers, and qk normalization.

For compute-constrained runs, we applied Gaussian smoothing with a window size of 301 (vs. 10 in Chinchilla), reducing variance by $\sim$13$\times$. This substantially improved fitting stability at the cost of a mild bias, corresponding to a $\sim$40-step lag. Learning rates were set to $2\mathrm{e}{-4}$ for models $<$8B and $1.25\mathrm{e}{-4}$ for models $>$8B, with a cosine decay schedule. To reuse prior epoch runs and collect stable data points, we employed the Warmup-Stable schedule \citep{hu2024minicpm} with peak learning rate $2\mathrm{e}{-4}$. 

All models were trained with sequence length 2048. Batch size was scaled with model size: 256 for $\leq$1.5B, 512 for 1.5B–5B, and 1024 for $>$5B, the latter chosen for stability.


\begin{figure}[ht]
\centering
\includegraphics[width=1\textwidth]{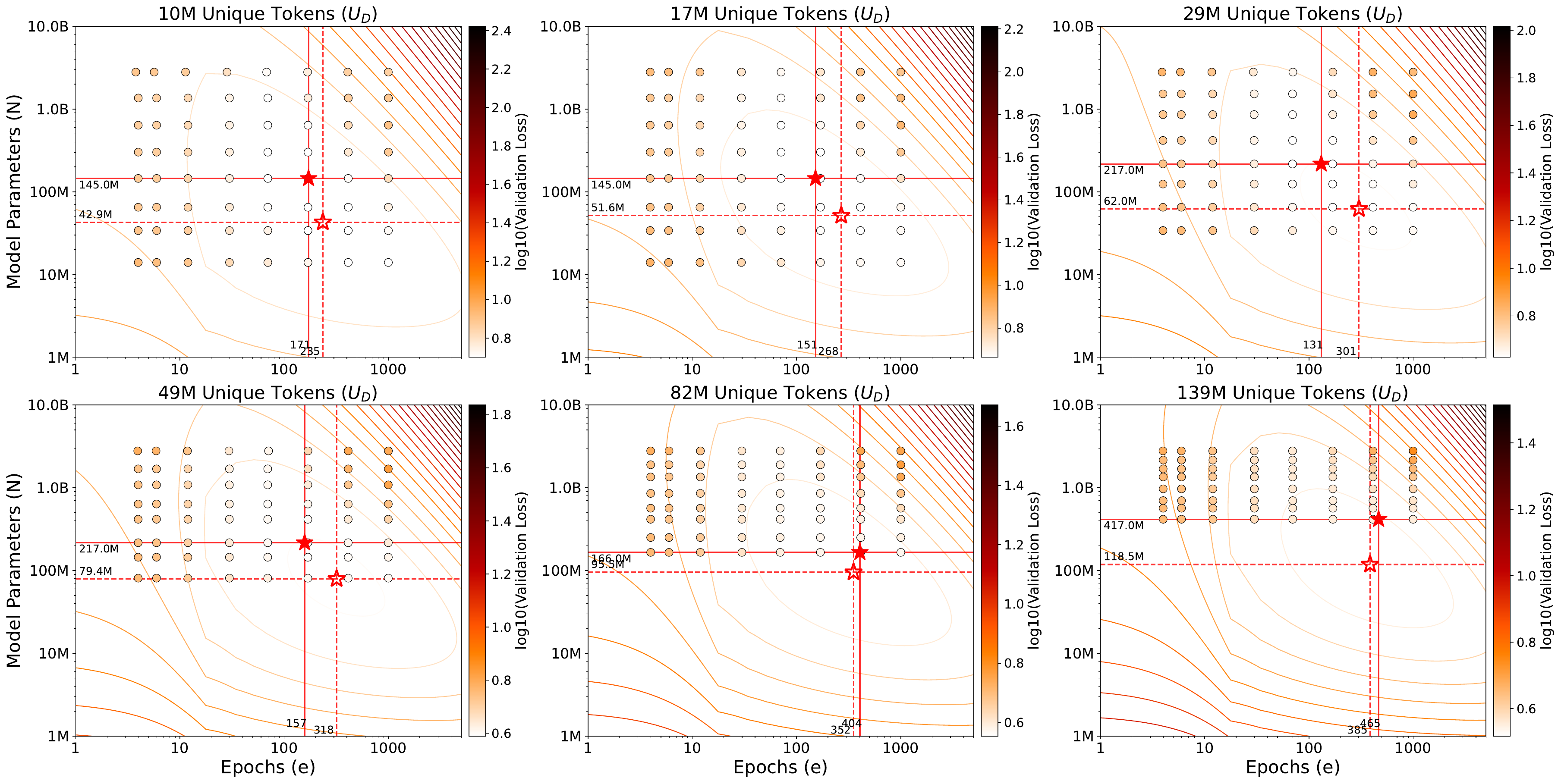}
\caption{\textbf{The contours predicted by Equation~\eqref{eq:fitted_dclaw} v.s. the real data points.} Equation~\eqref{eq:fitted_dclaw} tend to overshoot the epochs and under-estimate the model sizes.}
\label{fig:combined_scaling_law}
\end{figure}

\begin{figure}[ht]
\centering
\includegraphics[width=1\textwidth]{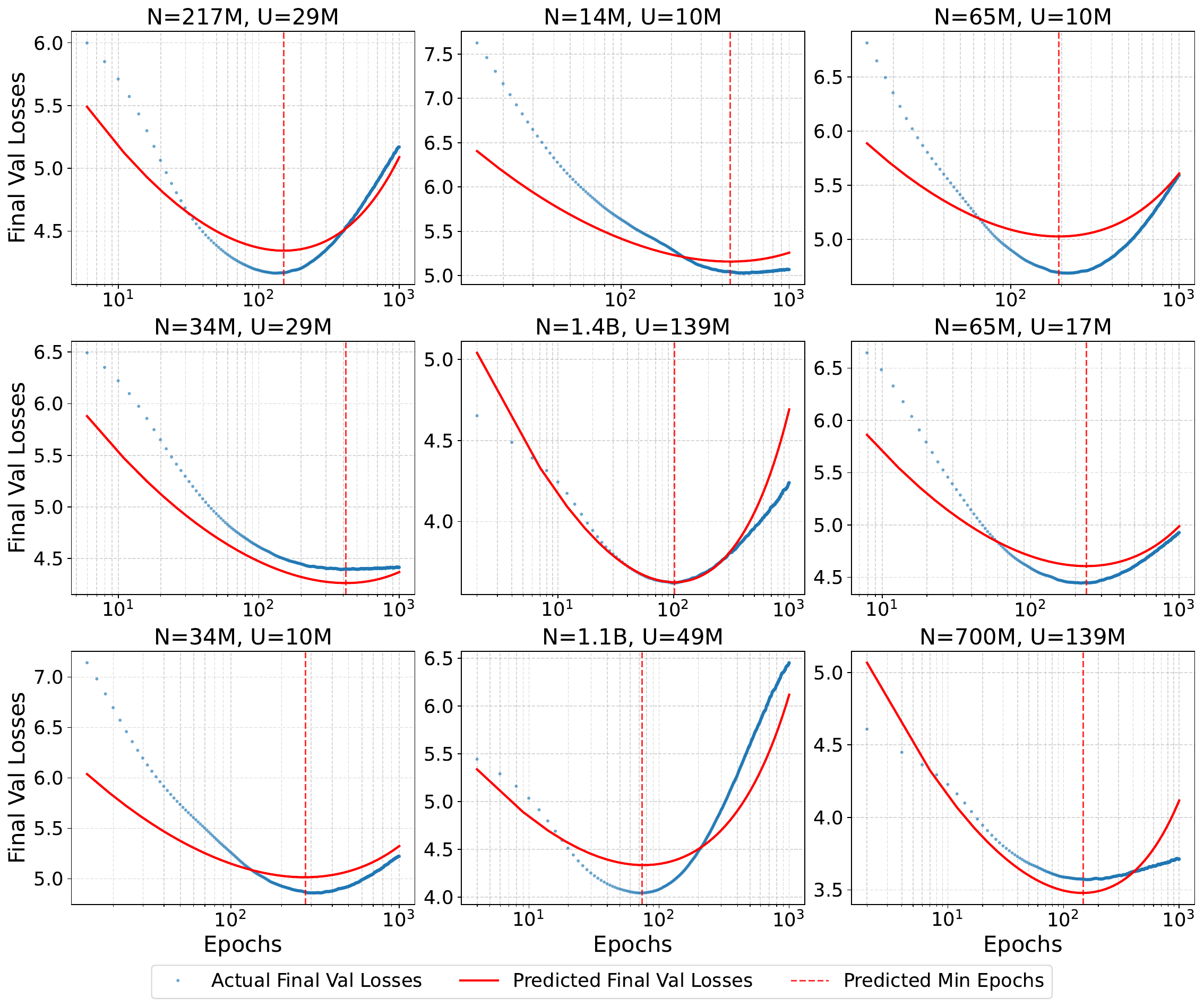}
\caption{The validation losses predicted by Equation~\eqref{eq:fitted_dclaw} v.s. the real validation losses.}
\label{fig:data_constrained_fitted_vs_actual}
\end{figure}

\section{Alternative Data-Constrained Formulas and Fitting Results}
\label{sec:data_constrained_alternative_formulas_and_fitting}

In this section we present the alternative data-constrained validation loss formulas, which are among the most effective ones we tried, but the losses (31.52 and 23.8 over 23145 data points) are still far from Equation~\eqref{eq:fitted_dclaw} (9.78 over 23145 data points).

\subsection{Additive Overfitting Term v1}

\begin{figure}[h]
\centering
\includegraphics[width=1\textwidth]{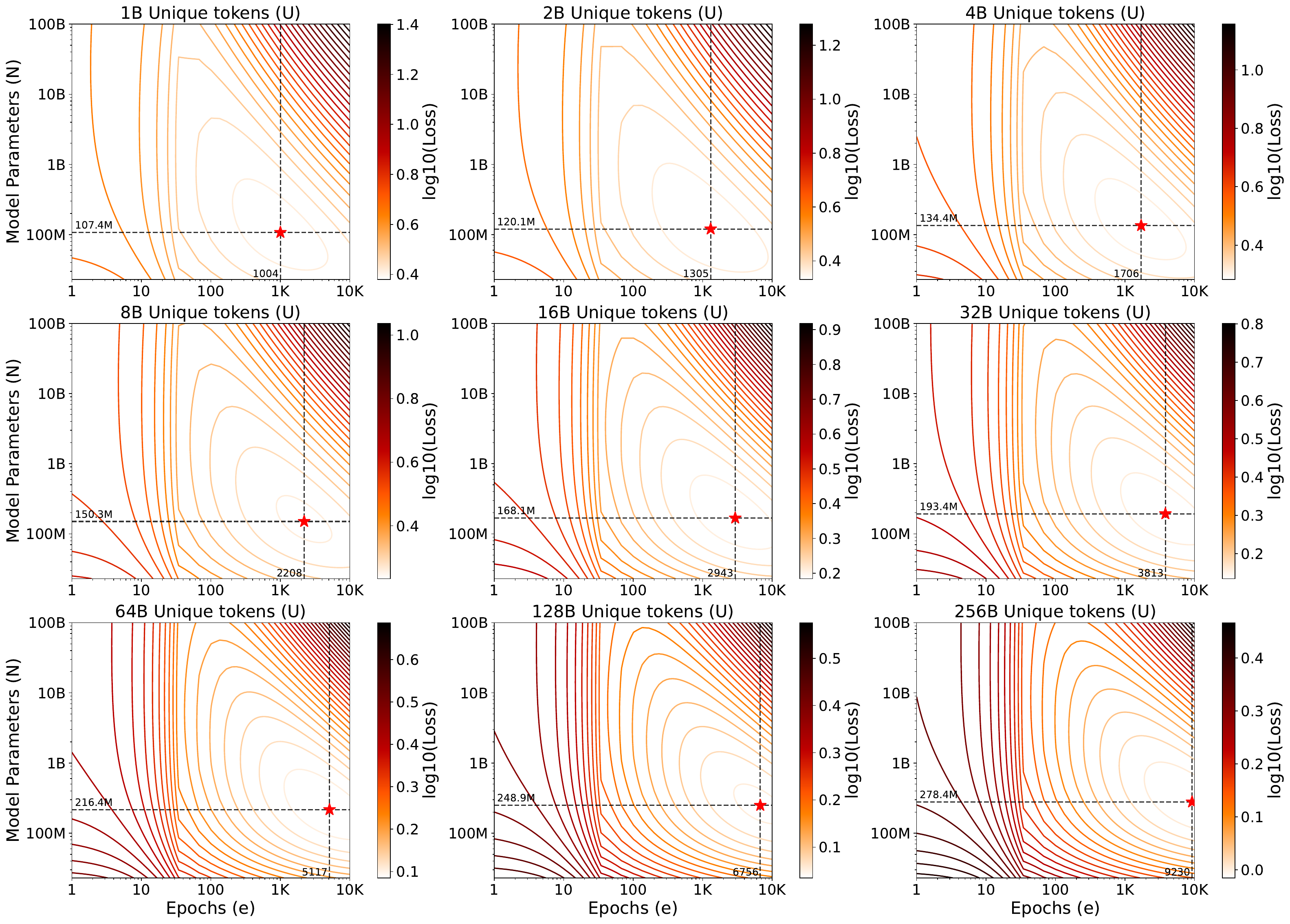}
\caption{The contours predicted by Equation~\eqref{ec:dclaw_additive_v1} and the optimal allocations.}
\label{fig:data_constrained_countour_v1}
\end{figure}

\begin{figure}[h]
\centering
\includegraphics[width=1\textwidth]{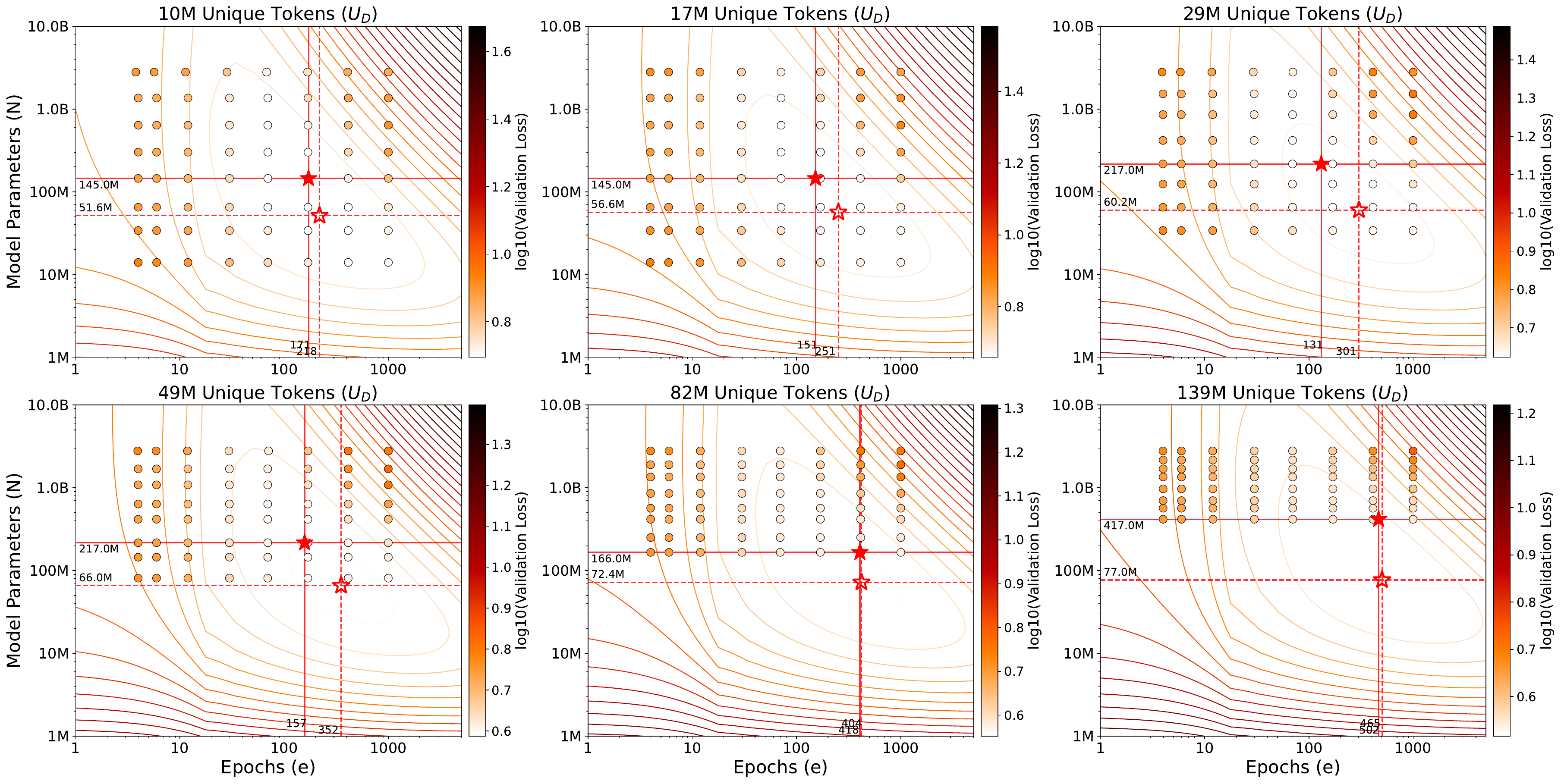}
\caption{The validation losses predicted by Equation~\eqref{ec:dclaw_additive_v1} v.s. the real validation losses.}
\label{fig:combined_scaling_law_additive_v1}
\end{figure}

Equation~\eqref{ec:dclaw_additive_v1} presents an additive formula breaking down the data-constrained scaling law into learning loss and overfitting penalty, with the fitted form in Equation~\eqref{eq:dclaw_additive_v1_fitted}. The predicted contours are presented in Figure~\ref{fig:data_constrained_countour_v1} and the contours v.s the actual data points are in Figure~\ref{fig:combined_scaling_law_additive_v1}.

\begin{equation}
    L(e, N, U_D) \triangleq \underbrace{E + \frac{A}{N^\alpha} + \frac{B}{(D'(e, U_D))^\beta}}_{\text{Learning Loss}} + \underbrace{\mu \left( \frac{N}{U_D} \right)^\delta \left( \log(\text{max}(1, e)) \right)^\gamma}_{\text{Overfitting Penalty}},
    \label{ec:dclaw_additive_v1}
\end{equation}

\begin{equation}
    \text{where} \quad D'(e, U_D) = U_D \left( 1 + R_D^* \left( 1 - exp(\frac{-\text{max}(0, e-1)}{R_D^*}) \right) \right),
\end{equation}

\begin{equation}
    D'(e, U_D) \approx U_D \cdot \text{max}(1, e), \text{as} \quad R_D^* \quad \text{is very large.}
\end{equation}

\begin{equation}
L(e, N, U_D) \approx \frac{145962.2}{N^{0.73}} + \frac{61.1}{\left[ U_D \cdot e \right]^{0.13}} +58 \times 10^{-4} \left( \frac{N}{U_D} \right)^{0.43} \left( \log(\text{max}(1, e)) \right)^{4.49}
\label{eq:dclaw_additive_v1_fitted}
\end{equation}

\subsection{Additive Overfitting Term v2}

\begin{figure}[h]
\centering
\includegraphics[width=1\textwidth]{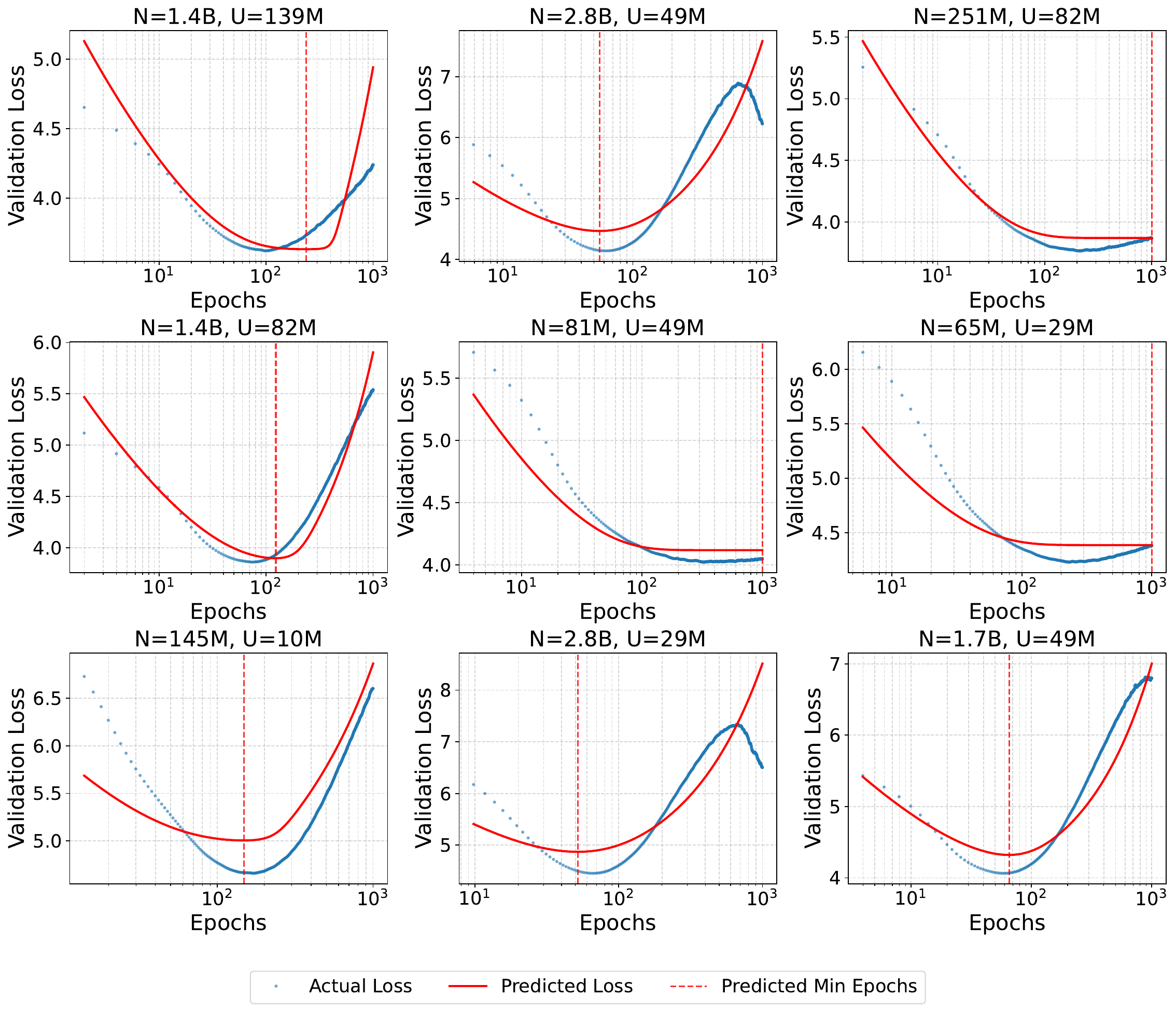}
\caption{The contours predicted by Equation~\eqref{ec:dclaw_additive_v2} v.s. the real data points.}
\label{fig:data_constrained_fitted_vs_actual_v2}
\end{figure}

\begin{figure}[h]
\centering
\includegraphics[width=1\textwidth]{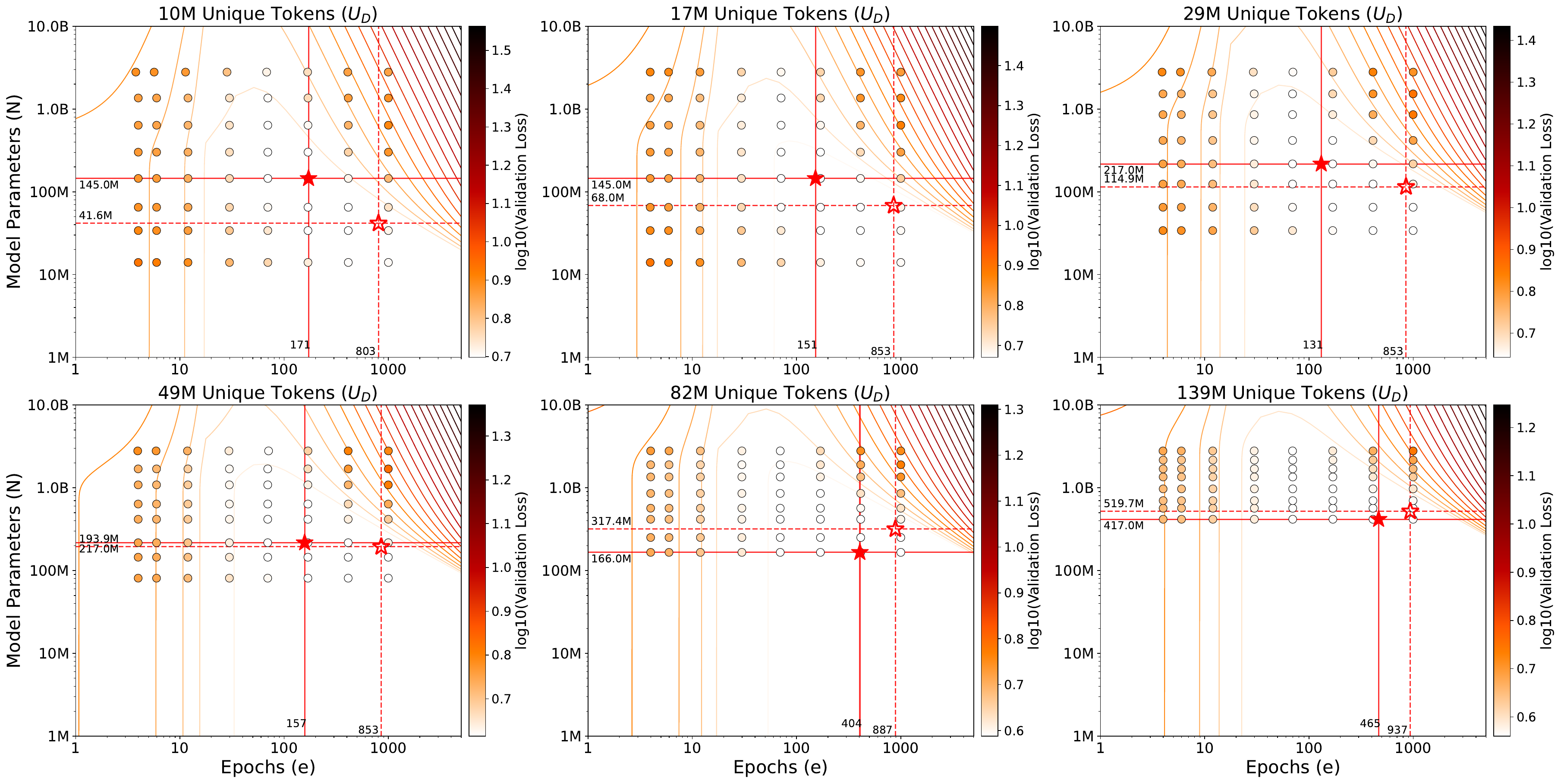}
\caption{The validation losses predicted by Equation~\eqref{ec:dclaw_additive_v2} v.s. the real validation losses.}
\label{fig:combined_scaling_law_additive_v2}
\end{figure}

Similarly, Equation~\eqref{ec:dclaw_additive_v2} presents an additive formula breaking down the data-constrained scaling law into learning loss and a more complicated overfitting penalty (after trials), with the fitted form in Equation~\eqref{eq:dclaw_additive_v2_fitted}. The predicted contours are presented in Figure~\ref{fig:data_constrained_fitted_vs_actual_v2} and the contours v.s the actual data points are in Figure~\ref{fig:combined_scaling_law_additive_v2}.

\begin{equation}
\begin{split}
L(e, N, U_D) &= \underbrace{E + \frac{A}{N^{\alpha}} + \frac{B}{D'^{\beta}}}_{\text{Learning Loss}} + \underbrace{\mu \left(\frac{N}{D'}\right)^{\delta} \left[ \text{softplus}\left( \frac{e - \kappa \left(\frac{U_D}{N}\right)^{\eta}}{\tau} \right) \right]^{\gamma}}_{\text{Overfitting Penalty}} \\
\text{where} \quad D'(e, U_D) &= U_D \left(1 + R_D^* \left(1 - \exp\left(-\frac{e-1}{R_D^*}\right)\right)\right) \\
\text{and} \quad \text{softplus}(x) &= \log(1 + e^x)
\end{split}
\label{ec:dclaw_additive_v2}
\end{equation}

\begin{equation}
\begin{split}
L(e, N, U_D) &\approx 9.505 \times 10^{-66} + \frac{2.738}{N^{1.240}} + \frac{53.58}{D'^{0.1207}} \\
&\quad+ 0.1610 \left(\frac{N}{D'}\right)^{0.3073} \left[ \text{softplus}\left( \frac{e - 12642 \left(\frac{U_D}{N}\right)^{1.486}}{26.56} \right) \right]^{0.8106} \\
\text{where} \quad D'(e, U_D) &= U_D \left(1 + 33.62 \left(1 - \exp\left(-\frac{e-1}{33.62}\right)\right)\right)
\end{split}
\label{eq:dclaw_additive_v2_fitted}
\end{equation}

\begin{figure}[h]
\centering
\includegraphics[width=1\textwidth]{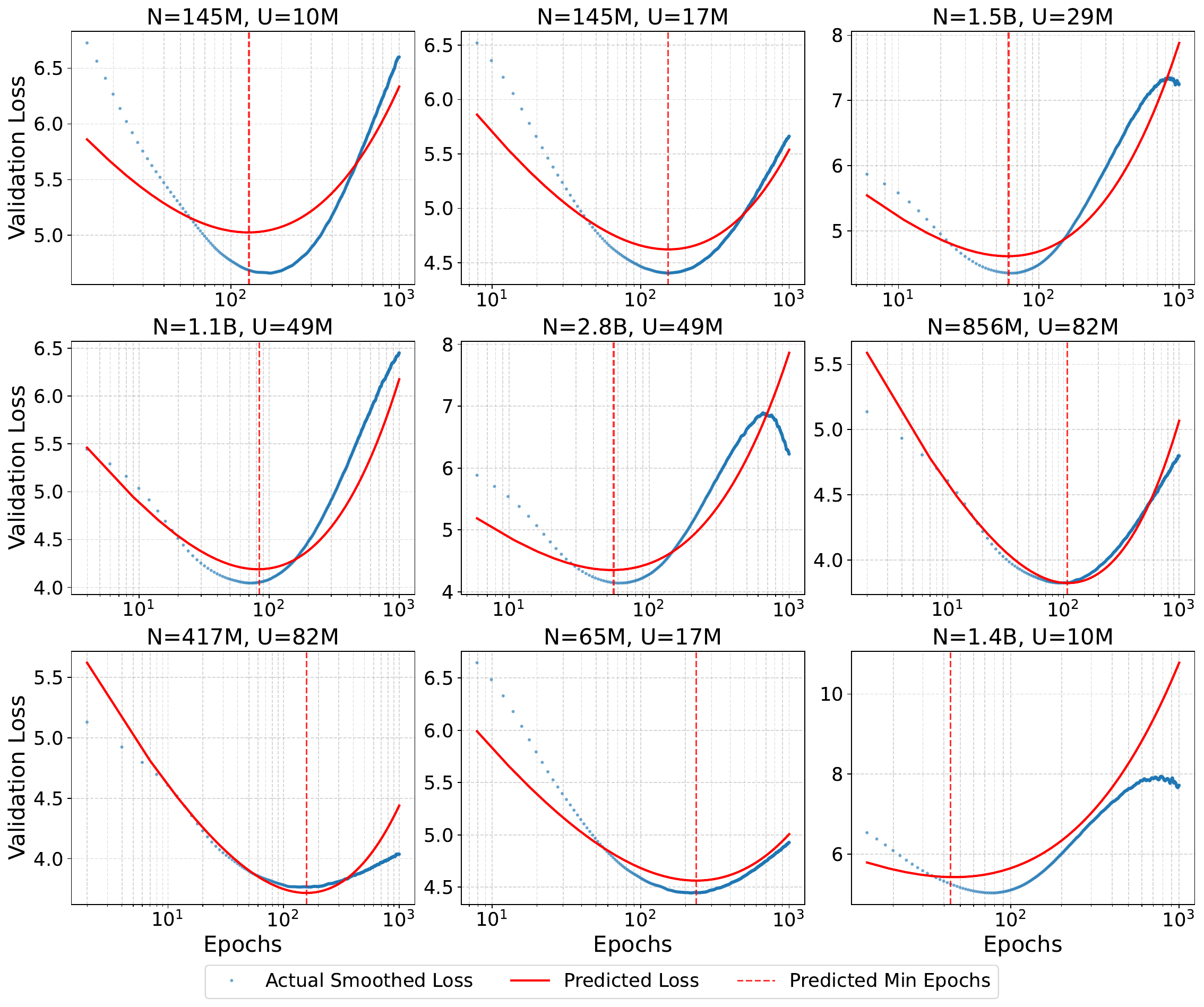}
\caption{The contours predicted by Equation~\eqref{ec:dclaw_additive_v1} v.s. the real data points.}
\label{fig:log_penalty_fitted_vs_actual_additive_v1}
\end{figure}

\begin{figure}[h]
\centering
\includegraphics[width=1\textwidth]{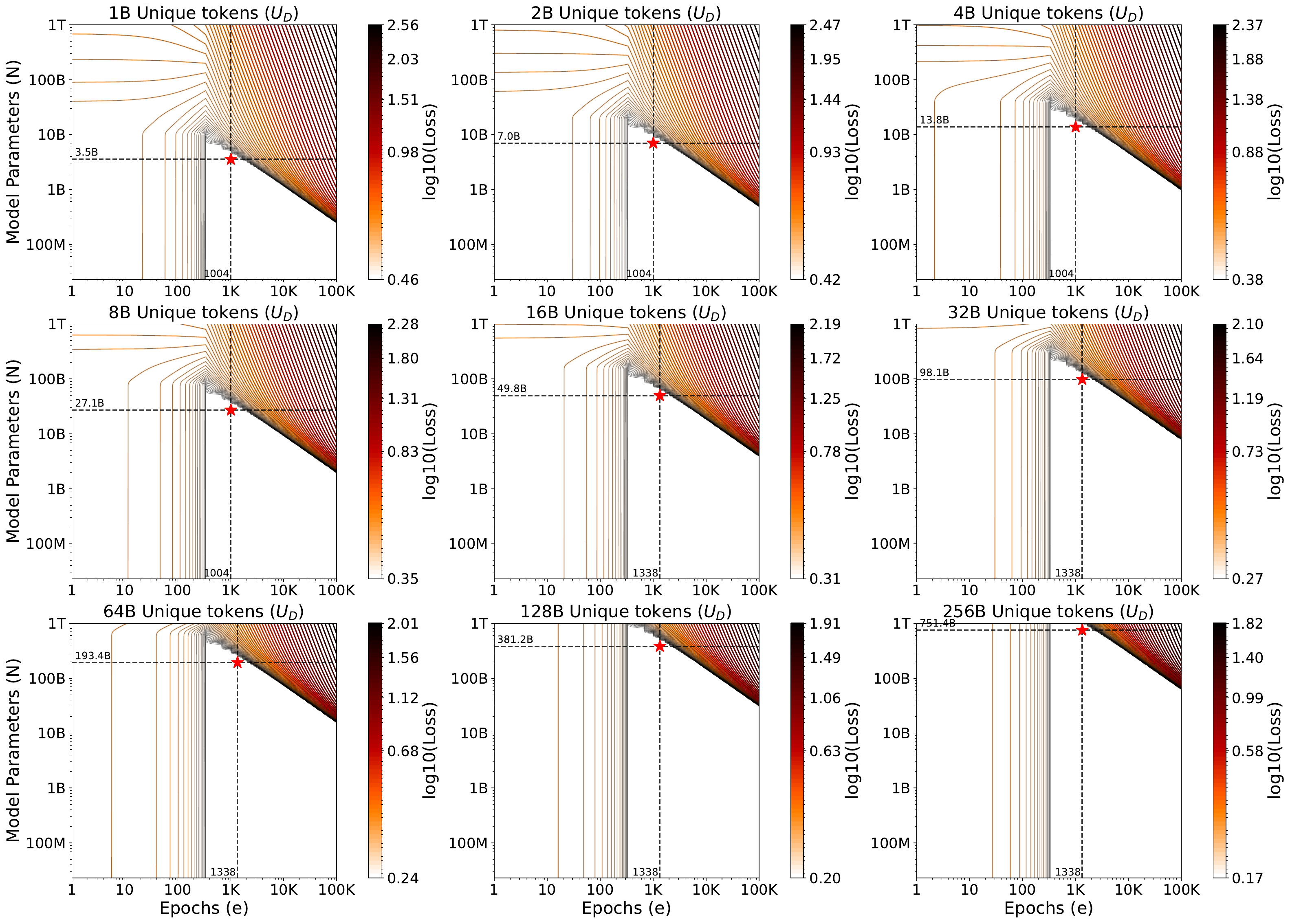}
\caption{The contours predicted by Equation~\eqref{ec:dclaw_additive_v2} and the optimal allocations.}
\label{fig:data_constrained_contour_additive_v2}
\end{figure}

\begin{table}[h]
\centering
\caption{\textbf{The FLOPs and Tokens allocation predicted by approach 2 and 3.} Similar to \citep{hoffmann2022training}, the loss fitting approach under-estimates $N \rightarrow N_{opt}$ for very large models.}
\label{tab:quokka_approach1n2}
\begin{tabular}{rrrrr}
\toprule
           & \multicolumn{2}{c}{Approach 1} & \multicolumn{2}{c}{Approach 2} \\ \midrule
Parameters & FLOPs           & Tokens       & FLOPs          & Tokens        \\ \midrule
400 M       & 9.46e+19        & 39.3 B        & 1.06e+20       & 44.3 B         \\
1 B         & 5.62e+20        & 93.5 B        & 6.68e+20       & 111.2 B        \\
10 B        & 4.96e+22        & 825.2 B       & 6.74e+22       & 1.1 T          \\
67 B        & 2.01e+24        & 5.0 T         & 3.05e+24       & 7.6 T          \\
175 B       & 1.30e+25        & 12.4 T        & 2.09e+25       & 19.9 T         \\
280 B       & 3.24e+25        & 19.3 T        & 5.35e+25       & 31.8 T         \\
520 B       & 1.08e+26        & 34.6 T        & 1.85e+26       & 59.3 T         \\
1 T         & 3.86e+26        & 64.2 T        & 6.86e+26       & 114.3 T        \\
10 T        & 3.41e+28        & 566.4 T       & 6.93e+28       & 1153.9 T       \\ \bottomrule
\end{tabular}
\end{table}

\begin{table}[t]
\footnotesize
\centering
\caption{The model arch details for all models trained in this work. All models used archs described in \S \ref{sec:imp_details}.}
\label{tab:model_params}
\begin{tabular}{cccccc}
\toprule
Parameters (million) & d\_model & ffw\_size & kv\_size & n\_heads & n\_layers \\ \midrule
1                    & 128     & 512      & 32      & 4       & 3        \\
2                    & 224     & 896      & 32      & 7       & 4        \\
5                    & 288     & 1,152    & 32      & 7       & 5        \\
7                    & 320     & 1,280    & 32      & 10      & 6        \\
14                   & 448     & 1,792    & 32      & 7       & 6        \\
25                   & 512     & 2,048    & 64      & 8       & 8        \\
36                   & 576     & 2,304    & 64      & 9       & 9        \\
49                   & 640     & 2,560    & 64      & 10      & 10       \\
64                   & 640     & 2,560    & 64      & 10      & 13       \\
79                   & 640     & 2,560    & 64      & 10      & 16       \\
85                   & 768     & 3,072    & 64      & 12      & 12       \\
106                  & 768     & 3,072    & 64      & 12      & 15       \\
127                  & 768     & 3,072    & 64      & 12      & 18       \\
135                  & 896     & 3,584    & 64      & 14      & 14       \\
154                  & 896     & 3,584    & 64      & 14      & 16       \\
173                  & 896     & 3,584    & 64      & 14      & 18       \\
201                  & 1,024   & 4,096    & 64      & 16      & 16       \\
226                  & 1,024   & 4,096    & 64      & 16      & 18       \\
252                  & 1,024   & 4,096    & 64      & 16      & 20       \\
354                  & 1,280   & 5,120    & 128     & 10      & 18       \\
413                  & 1,280   & 5,120    & 128     & 10      & 21       \\
428                  & 1,408   & 5,632    & 128     & 11      & 18       \\
472                  & 1,280   & 5,120    & 128     & 10      & 24       \\
500                  & 1,408   & 5,632    & 128     & 11      & 21       \\
538                  & 1,536   & 6,144    & 128     & 12      & 19       \\
571                  & 1,408   & 5,632    & 128     & 11      & 24       \\
623                  & 1,536   & 6,144    & 128     & 12      & 22       \\
708                  & 1,536   & 6,144    & 128     & 12      & 25       \\
771                  & 1,792   & 7,168    & 128     & 14      & 20       \\
886                  & 1,792   & 7,168    & 128     & 14      & 23       \\
1,002                & 1,792   & 7,168    & 128     & 14      & 26       \\
1,107                & 2,048   & 8,192    & 128     & 16      & 22       \\
1,250                & 2,176   & 8,704    & 128     & 17      & 22       \\
1,258                & 2,048   & 8,192    & 128     & 16      & 25       \\
1,409                & 2,048   & 8,192    & 128     & 16      & 28       \\
1,420                & 2,176   & 8,704    & 128     & 17      & 25       \\
1,529                & 2,304   & 9,216    & 128     & 18      & 24       \\
1,591                & 2,176   & 8,704    & 128     & 17      & 28       \\
1,784                & 2,304   & 9,216    & 128     & 18      & 28       \\
2,038                & 2,304   & 9,216    & 128     & 18      & 32       \\
2,045                & 2,560   & 10,240   & 128     & 20      & 26       \\
2,359                & 2,560   & 10,240   & 128     & 20      & 30       \\
2,674                & 2,560   & 10,240   & 128     & 20      & 34       \\
3,121                & 2,688   & 10,752   & 128     & 21      & 36       \\
3,426                & 2,816   & 11,264   & 128     & 22      & 36       \\
3,744                & 2,944   & 11,776   & 128     & 23      & 36       \\
4,077                & 3,072   & 12,288   & 128     & 24      & 36       \\
6,166                & 3,584   & 14,336   & 128     & 28      & 40       \\
8,456                & 4,096   & 16,384   & 128     & 32      & 42       \\
10,682               & 4,352   & 17,408   & 128     & 32      & 47       \\
11,211               & 4,608   & 18,432   & 128     & 36      & 44       \\
11,976               & 4,608   & 18,432   & 128     & 32      & 47       \\
13,343               & 4,864   & 19,456   & 128     & 32      & 47       \\
14,653               & 4,992   & 19,968   & 128     & 32      & 49       \\
14,785               & 5,120   & 20,480   & 128     & 40      & 47       \\ \bottomrule
\end{tabular}
\end{table}

\end{document}